\documentclass{article}

\usepackage{PRIMEarxiv}

\usepackage[utf8]{inputenc} 
\usepackage[T1]{fontenc}    
\usepackage{hyperref}       
\usepackage{url}            
\usepackage{booktabs}       
\usepackage{amsfonts}       
\usepackage{nicefrac}       
\usepackage{microtype}      
\usepackage{lipsum}
\usepackage{fancyhdr}       
\usepackage{graphicx}       
\graphicspath{{media/}}     
\usepackage{multirow}
\usepackage{amsmath}
\usepackage{algorithm}
\usepackage{algorithmic}
\usepackage{tabularx}
\usepackage{float}
\usepackage{pdflscape}    
\usepackage{longtable}
\usepackage{array}
\usepackage{ltablex}      
\usepackage{ragged2e}
\usepackage{threeparttable}  
\usepackage{caption}         

\title{CardioForest: An Explainable Ensemble Learning Model for Automatic Wide QRS Complex Tachycardia Diagnosis from ECG
\thanks{\textit{{These are corresponding authors.}}\\ 
\textit{† These authors contributed equally.}} 
}

\author{
  Vaskar Chakma † \\
  School of Artificial Intelligence and Computer Science \\
  Nantong University \\
  Jiangsu, China \\
  \texttt{vaskarchakma@stmail.ntu.edu.cn} \\
   \And
  Ju Xiaolin † \\
  School of Artificial Intelligence and Computer Science \\
  Nantong University \\
  Jiangsu, China \\
  \texttt{ju.xl@ntu.edu.cn} \\
  \And
  Heling Cao \\
  College of Information Science and Engineering \\
  Henan University of Technology \\
  Zhengzhou, China \\
  \texttt{caohl@haut.edu.cn} \\
  \And
  Xue Feng, Ji Xiaodong, Pan Haiyan *  \\
  Affiliated Hospital of Nantong University \\
  Jiangsu, China \\
  \texttt{\{xuefengtdfy, 5201199, dr.phy\}@ntu.edu.cn} \\
  \And
  Gao Zhan * \\
  School of Artificial Intelligence and Computer Science \\
  Nantong University \\
  Jiangsu, China \\
  \texttt{gaozhan@ntu.edu.cn} \\
}

\begin{document}
\maketitle

\begin{abstract}
This study aims to develop and evaluate an ensemble machine learning-based framework for the automatic detection of Wide QRS Complex Tachycardia (WCT) from ECG signals, emphasizing diagnostic accuracy and interpretability using Explainable AI. The proposed system integrates ensemble learning techniques, i.e., an optimized Random Forest known as CardioForest, and models like XGBoost and LightGBM. The models were trained and tested on ECG data from the publicly available MIMIC-IV dataset. The testing was carried out with the assistance of accuracy, balanced accuracy, precision, recall, F1 score, ROC-AUC, and error rate (RMSE, MAE) measures. In addition, SHAP (SHapley Additive exPlanations) was used to ascertain model explainability and clinical relevance. The CardioForest model performed best on all metrics, achieving a test accuracy of 95.19\% ($\pm$0.33\%), a balanced accuracy of 88.76\% ($\pm$0.79\%), precision of 95.26\%, recall of 78.42\%, and ROC-AUC of 0.8886. SHAP analysis confirmed the model's ability to rank the most relevant ECG features, such as QRS duration, in accordance with clinical intuitions, thereby fostering trust and usability in clinical practice. The findings recognize CardioForest as an extremely dependable and interpretable WCT detection model. Being able to offer accurate predictions and transparency through explainability makes it a valuable tool to help cardiologists make timely and well-informed diagnoses, especially for high-stakes and emergency scenarios.
\end{abstract}


\section{Introduction}
\label{sec:introduction}
Wide QRS Complex Tachycardia (WCT) is a severe and potentially lethal cardiac condition characterized by an exceedingly rapid heartbeat in combination with a widened QRS complex on the electrocardiogram (ECG) \cite{alblaihed2022wide, song2022electrocardiographic, fayyazifar2023novel}. Normally, the QRS complex—a short, spiky waveform—registers the process of ventricular depolarization, whereby the ventricles of the heart contract and effectively pump blood \cite{kurl2012duration, silvetti2023pivotal}. A regular narrow QRS complex indicates typical electrical conduction through the heart's normal pathways \cite{hampton2024ecg, badura2024primary}. However, if the QRS complex is wide, then this is an indication of a disruption in electrical propagation \cite{jastrzebski2012comparison}, typically as a result of underlying structural disease, electrolyte imbalance, or an inherited disorder. 
\begin{figure}
\centering
\includegraphics[scale=0.32]{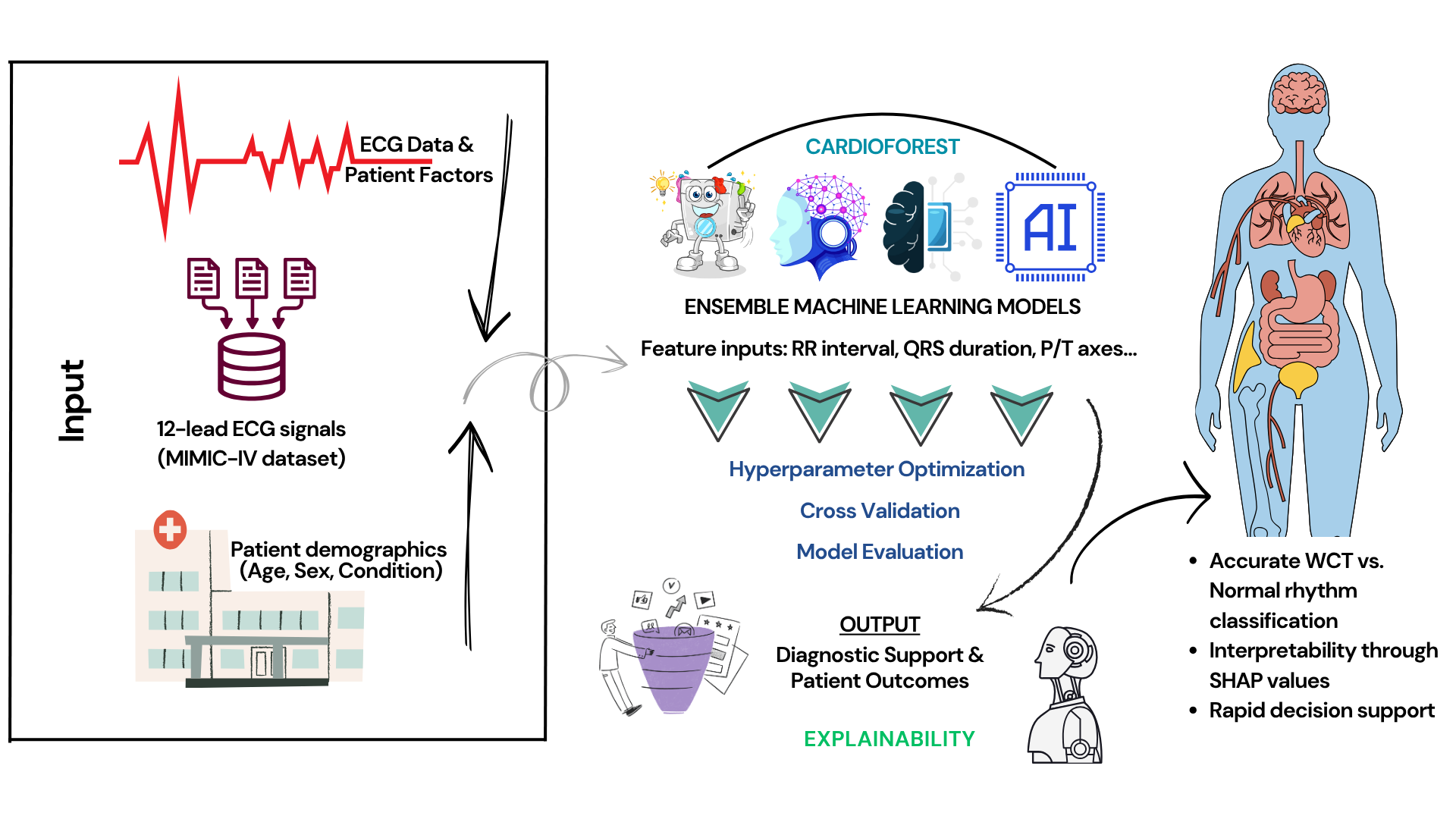}
\caption{An overview of the WCT prediction system using the MIMIC-IV ECG database, featuring preprocessing, ensemble machine learning models, cross-validation, and final prediction.}
\label{fig 1}
\end{figure}
Untreated, WCT can significantly weaken the heart's function to circulate blood effectively, causing symptoms that range from palpitations, dizziness, and chest pain to, in extreme cases, sudden cardiac arrest \cite{osadchii2017role, tse2016mechanisms}. As such, the early and correct diagnosis of WCT is not only critical—it is a matter of life and death. Diagnosis of WCT has traditionally depended to a large extent on manual ECG interpretation by experienced cardiologists. While still the gold standard, this process is time-consuming, labor-intensive, and subject to considerable variability \cite{adib2023generating, yang2025application}. Individual cardiologists may disagree in borderline or uncertain cases, postponing diagnosis and treatment \cite{adewole2022expert, meyer2021patient}. In high-pressure clinical environments where minutes matter, delays can be detrimental. As healthcare systems globally face rising demands, the demand for faster, more accurate diagnostic support that augments, rather than replaces, clinical judgment is pressing. In the past few years, Artificial Intelligence (AI)-driven models have been demonstrated to achieve stellar performance in ECG interpretation with accuracy and speed \cite{muzammil2024artificial}. Among them, deep learning methods—particularly Convolutional Neural Networks (CNNs)—have worked incredibly well in identifying complex patterns within ECG signals that are not easily visible to the naked eye \cite{abubaker2022detection, ahmed2023classifying, ramkumar2021graphical}. However, for all the high-accuracy deep learning models claim, they tend to behave like "black boxes" with little description of decision-making. This absence of transparency has been a significant barrier to clinical adoption because cardiologists and clinicians need not only accuracy but also transparency to trust AI recommendations. For AI to be successfully integrated into clinical practice, especially in life-critical conditions like WCT, interpretability is equally as important as accuracy \cite{brisk2023towards}. Cardiologists must understand the rationale for AI predictions—seeing not just the output, but the supporting evidence, e.g., what ECG features led to a specific classification. Without this transparency, clinicians will remain unconvinced and reluctant to trust AI for decision-making, especially when a patient's life hangs in the balance. Given these challenges, this research introduces a novel solution: \textbf{CardioForest}, an interpretable ensemble-based AI model for WCT detection. Based primarily on Random Forest architecture—augmented with techniques such as XGBoost, LightGBM, and Gradient Boosting—CardioForest leverages the strengths of ensemble machine learning to achieve both high diagnostic precision and clear interpretability \cite{nissa2024technical}. Unlike deep neural networks, Random Forest models are natively explainable through feature importance rankings and decision-tree visualization. This dual advantage ensures clinicians can rely on and interpret the model outputs. Our experiment design is focused not only on the evaluation of diagnostic accuracy but also on increasing clinical trust by employing explainable AI (XAI) methods, for example, SHAP (SHapley Additive exPlanations) \cite{lundberg2017unified} values and feature attribution analysis \cite{sadeghi2024review}. CardioForest bridges the crucial gap between AI’s computational power and the clinician’s need for interpretability by highlighting which features most significantly influenced the model's decisions. Preliminary results suggest that CardioForest outperforms traditional manual approaches and competes favorably with state-of-the-art deep learning models while offering superior transparency, an essential quality for clinical adoption. For all these developments, we acknowledge that challenges remain. Future research should also explore continuous learning frameworks, where AI models learn incrementally from new data, thereby improving their diagnostic acumen over time without compromising explainability. Figure \ref{fig 1} presents \textit{CardioForest} as a pioneering WCT detection solution—diagnostic performance coupled with interpretability \cite{chakma2025cardioforest}, a union necessary for real-world clinical practice. By providing cardiologists with speedy, interpretable, and reliable AI support, we hope to enhance cardiac diagnosis, reduce diagnostic latency, and ultimately save more lives. Looking ahead, we envision the expansion of explainable ensemble model applications beyond WCT toward general arrhythmia detection with the inclusion of real-time ECG monitoring for preemptive cardiac health management.

\begin{table}
\caption{List of Abbreviations}
\label{tab:abbreviations}
\centering
\begin{tabular}{@{}ll@{}}
\hline
\textbf{Abbreviation} & \textbf{Definition} \\
\hline
AI & Artificial Intelligence \\
AUC & Area Under the Curve \\
CNN & Convolutional Neural Network \\
ECG & Electrocardiogram \\
GBM & Gradient Boosting Machine \\
LGBM & Light Gradient Boosting Machine \\
MIMIC & Medical Information Mart for Intensive Care \\
PCA & Principal Component Analysis \\
RF & Random Forest \\
RMSE & Root Mean Square Error \\
ROC & Receiver Operating Characteristic \\
SHAP & SHapley Additive exPlanations \\
ST & Standard Deviation \\
WCT & Wide QRS Complex Tachycardia \\
WFDB & WaveForm DataBase \\
XAI & Explainable Artificial Intelligence \\
XGBoost & Extreme Gradient Boosting \\
\hline
\end{tabular}
\end{table}

\section{Related Works}
Accurate and timely prediction of Wide Complex Tachycardia (WCT) remains a major focus in cardiovascular research, driven by the need to distinguish between ventricular tachycardia (VT) and supraventricular tachycardia (SVT) with aberrant conduction. Machine learning (ML) and deep learning (DL) techniques have gained prominence in this domain \cite{chakma2025machine}, offering new avenues for improved diagnostic performance compared to traditional criteria-based methods.

Li et al. \cite{Li2024} proposed a Gradient Boosting Machine (GBM) model for differentiating VT from SVT using surface ECG features. Their approach leveraged a rich set of ECG-derived parameters, leading to an impressive classification performance with an overall accuracy of 91.2\%, sensitivity of 89.5\%, specificity of 92.8\%, and an area under the ROC curve (AUC) of 0.94. This study highlighted the importance of carefully selected ECG features in enhancing machine learning model performance. Building on the trend of AI-driven diagnosis, Chow et al. \cite{Chow2024} developed a specialized AI model to interpret WCT directly from ECGs. Their system, designed with clinical applicability in mind, demonstrated an overall accuracy of 93\%, with sensitivity and specificity exceeding 91\%. This work showcased the potential of deep learning models in outperforming traditional rule-based algorithms for arrhythmia classification, particularly for ambiguous WCT cases.

Focusing on high-risk populations, Bhattacharya et al. \cite{Bhattacharya2021} introduced the HCM-VAr-Risk model, which applies machine learning techniques to predict ventricular arrhythmias in patients with hypertrophic cardiomyopathy (HCM). Their model achieved a C-index of 0.83, reflecting strong discriminative ability. The study underscored the utility of ML for risk stratification in structurally abnormal hearts, offering a more individualized approach to arrhythmia prediction. Hong et al. \cite{Hong2020} provided a broader perspective by conducting a systematic review of deep learning applications for ECG analysis, including arrhythmia detection and classification tasks. The review covered a range of architectures, such as convolutional neural networks (CNNs), recurrent neural networks (RNNs), and hybrid models, illustrating the high accuracy and generalizability of DL models when trained on large, diverse ECG datasets. Their findings support the growing consensus that deep learning can significantly enhance the detection of complex arrhythmias, including WCT.

Addressing diagnostic challenges from a different angle, May et al. \cite{May2024} introduced the QRS Polarity Shift (QRS-PS) method, which focuses on changes in QRS polarity between baseline ECGs and WCT episodes. By simplifying the interpretation of polarity shifts, their algorithm achieved AUC values ranging from 0.90 to 0.93. This technique provides a pragmatic and explainable tool that can be readily integrated into clinical workflows, assisting clinicians in making rapid and accurate diagnoses. Machine learning classification models have also shown remarkable potential in SVT detection. Howladar and Sahoo \cite{howladar2021} developed a decision-tree-based model specifically tailored for SVT identification. Their model attained a striking 97\% accuracy, demonstrating the effectiveness of even relatively simple ML algorithms when paired with relevant feature selection. Deep learning models have further pushed the boundaries of arrhythmia prediction. Rajpurkar et al. \cite{Rajpurkar2017} designed a CNN-based model, trained on a large annotated ECG dataset, that achieved recall and precision rates exceeding those of board-certified cardiologists. Their work set a new benchmark for DL-based arrhythmia detection and provided strong evidence for adopting AI-assisted ECG interpretation tools in clinical practice.

\begin{table}
\caption{Summary of Diverse Studies on WCT Prediction}
\label{tab:related_works}
\centering
\begin{tabular}{@{} p{2.5cm} p{3.2cm} p{4.5cm} p{3cm} p{1.7cm} @ {}}
\hline
\textbf{Study} & \textbf{Model/Method} & \textbf{Key Features} & \textbf{Performance Metrics} & \textbf{References} \\
\hline
Li et al. (2024)  & Gradient Boosting Machine (GBM) & Differentiates VT from SVT using surface ECG features & Accuracy: 91.2\%, Sensitivity: 89.5\%, Specificity: 92.8\%, AUC: 0.94 & \cite{Li2024}  \\
Chow et al. (2024)  & Artificial Intelligence (AI) Model & AI algorithm interpreting WCT ECGs & Accuracy: 93\%, Sensitivity: 91.9\%, Specificity: 93.4\% & \cite{Chow2024}\\
Bhattacharya et al. (2021) & HCM-VAr-Risk Model & ML model for ventricular arrhythmias in HCM patients & Sensitivity: 73\%, Specificity: 76\%, C-index: 0.83 & \cite{Bhattacharya2021}\\
Hong et al. (2020) & Deep Learning Review & Systematic review of DL methods for ECG data & Various models achieving high accuracy and AUC & \cite{Hong2020} \\
May et al. (2024) & QRS Polarity Shift (QRS-PS) & Algorithm based on QRS polarity shifts between WCT and baseline ECGs & AUC: 0.90–0.93 & \cite{May2024} \\
Howladar \& Sahoo (2021) & Classification Model & ML-based SVT detection \& decision-tree-based model for SVT classification & Accuracy: 97\% & \cite{howladar2021} \\
Rajpurkar et al. (2017) & CNN-based Model & Deep learning model for arrhythmia detection & Exceeds average cardiologist performance in recall and precision & \cite{Rajpurkar2017} \\
Frausto-Avila et al. (2024) & Compact Neural Network & ANN with feature enhancement for arrhythmia classification & Accuracy: 97.36\% & \cite{fraustoavila2024} \\
\hline
\end{tabular}
\end{table}

In addition, Frausto-Avila et al. \cite{fraustoavila2024} presented a compact neural network architecture enhanced with advanced feature engineering techniques. Their model achieved an accuracy of 97.36\% in arrhythmia classification tasks, suggesting that lightweight models can maintain high predictive performance while offering advantages in computational efficiency, making them suitable for deployment in real-time or resource-constrained environments. The reviewed studies demonstrate that both machine learning and deep learning approaches have significantly advanced WCT prediction and arrhythmia classification. The diversity of methods—from feature-driven models like GBM and decision trees to sophisticated deep learning architectures like CNNs—reflects the rich potential of AI technologies to improve clinical outcomes \cite{vaskar3margins}. A detailed comparison of these related works, including their methodologies, key innovations, and achieved performance metrics, is presented in table~\ref{tab:related_works}.

\section{Diagnostic Data Resources}  
This study utilizes the MIMIC-IV-ECG \cite{BrianGow2023}
Module (a statistical summary of the dataset has been shown in table \ref{tab:data_sources}), a comprehensive database of diagnostic electrocardiogram (ECG) waveforms \cite{goldberger} integrated with the broader MIMIC-IV Clinical Database. The dataset contains approximately 800,000 ten-second-long 12-lead ECG recordings sampled at 500 Hz, collected from
nearly 160,000 unique patients. For computational efficiency while maintaining statistical validity,
we utilized a stratified random sample of recordings (see Section~\ref{sec:sampling_strategy}),
preserving the original class distribution (15.46\% WCT prevalence). Each electrocardiogram (ECG)
record is stored in the standard WaveForm DataBase (WFDB)\footnote{https://physionet.org/lightwave/}
format, which includes a header file (.hea) and a binary data file (.dat). The records are organized in a structured directory hierarchy based on the subject identifier, allowing for efficient data retrieval. For example, a subject with ID 10001725 would be stored under the path files/p1000/p10001725/, with each diagnostic study within a subdirectory labeled by a randomly generated study ID. Approximately 55\% of the ECGs in the dataset overlap with a hospital admission and 25\% with an emergency department visit, while the remaining records were collected outside traditional inpatient or emergency settings. This diversity in acquisition context allows for a wide range of use cases, from acute event analysis to routine monitoring assessments. However, it is important to note that the ECG timestamps are derived from the internal clock of the acquisition device and are not synchronized with the hospital's clinical information systems. 

\begin{table}
\caption{Statistics of the MIMIC-IV real-world datasets used in this paper}
\label{tab:data_sources}
\centering
\setlength{\tabcolsep}{20pt}
\begin{tabular}{p{120pt} p{50pt}}
\hline
\textbf{Metric} & \textbf{Value} \\
\hline
Total Records & 800,035 \\
Unique Subjects & 161,352 \\
Unique Studies & 800,035 \\
Unique Carts & 156 \\
Average RR Interval (ms) & 865.60 \\
Average QRS Duration (ms) & 108.51 \\
Average P Onset (ms) & 4,702.88 \\
Average P End (ms) & 8,745.08 \\
Average QRS Onset (ms) & 283.42 \\
Average QRS End (ms) & 391.66 \\
Average T End (ms) & 688.65 \\
Average P Axis (degrees) & 4,973.35 \\
Average QRS Axis (degrees) & 107.37 \\
Average T Axis (degrees) & 192.55 \\
\hline
\end{tabular}
\end{table}

Consequently, temporal alignment between the ECGs and clinical events in the MIMIC-IV database may require additional validation. Each ECG waveform is accompanied by machine-generated summary measurements \cite{BrianGow2023} stored in the machine\_measurements.csv file. These include standard parameters such as RR interval, QRS onset and end, and filter settings, along with textual machine-generated interpretation notes across columns report\_0 to report\_17. The accompanying data dictionary in machine\_measurements-\_data\_dictionary.csv describes the technical and clinical meaning of each column. Each record includes a subject\_id, study\_id, and ecg\_time, enabling direct linkage to clinical data in the MIMIC-IV hospital and emergency department modules.
Cardiologist interpretations are also available for over 600,000 ECG studies. These free-text reports are stored in the MIMIC-IV-Note module and are linked to the ECG waveforms via the waveform\_note\_links.csv file. Each entry in this linkage file includes the subject ID, study ID, waveform path, and a note\_id that can be used to retrieve the corresponding cardiologist report. A sequential integer (note\_seq) is also provided to determine the order of ECG collection for individual patients. This linkage enables researchers to perform comparative analyses between machine-generated and clinician-interpreted findings. To support large-scale analysis, key metadata from record\_list.csv, machine\_measurements.csv, and waveform\_note\_links.csv have been made available through Google BigQuery. This facilitates efficient querying and integration with other clinical tables in the MIMIC-IV ecosystem. As a practical illustration, using BigQuery, a researcher can identify a patient’s hospital admissions and correlate them with the timing of their ECGs, determine whether a given ECG occurred during a hospital stay, and check for the presence of associated cardiologist notes \cite{li2025icu}, \cite{yoon2024redefining}. For waveform visualization and signal processing, the dataset supports standard PhysioNet WFDB toolkits in Python, MATLAB, and C. Researchers can read and visualize ECG waveforms using the wfdb Python package \cite{sharma2023wfdb}. For instance, using wfdb.rdrecord() and wfdb.plot\_wfdb(), one can extract and display the raw signal for any given ECG study. This compatibility makes the dataset highly accessible for both signal processing and clinical informatics researchers.
Despite its richness, the dataset has some limitations. Notably, the ECG device timestamps may be inaccurate due to a lack of clock synchronization. Additionally, some ECGs were recorded outside the hospital or emergency department, limiting direct temporal correlation with clinical events \cite{goodwin2022timing}. Nonetheless, MIMIC-IV-ECG is invaluable for studying cardiac health, machine learning applications in ECG interpretation, and cross-modal linkage with comprehensive clinical records \cite{chakma2025cardioforest}.

\section{Data Preparation and Processing Pipeline}  
\subsection{Data Cleaning and Preprocessing Techniques} 
Duplicate entries were identified using \texttt{subject\_id}  and \texttt{study\_id}, ensuring each ECG was uniquely represented. Categorical variables (e.g., wct\_label) were encoded numerically using label encoding, and floating-point precision errors were truncated. This ensured compatibility with machine learning algorithms and improved computational efficiency. The complete preprocessing pipeline is detailed in Algorithm \ref{alg:preprocessing}. Pandas' duplicated() function \cite{gupta2024data} detected redundant records, which were subsequently removed. Post-cleaning verification confirmed the dataset retained its integrity, with zero duplicate records. From the cleaned dataset of 800,035 records, we extracted a stratified sample of records for analysis (see Section~\ref{sec:sampling_strategy}), ensuring
representative distribution across all clinical variables while maintaining computational tractability
for comprehensive cross-validation and explainability analyses. Biologically implausible values (e.g., negative RR intervals) were corrected using interpolation \cite{morelli2019analysis}, while extreme outliers were adjusted or removed. Visualizations like boxplots and histograms validated the corrections, showing normalized distributions for key features such as RR intervals and QRS durations. Timestamps (ecg\_time\_x and ecg\_time\_y) were converted to a uniform format using Python’s datetime module, ensuring consistency for time-series analysis. This step addressed discrepancies arising from unsynchronized ECG machine clocks. Categorical variables (e.g., wct\_label) were encoded numerically using label encoding, and floating-point precision errors were truncated. This ensured compatibility with machine learning algorithms and improved computational efficiency.

\begin{figure}
\centering
\includegraphics[scale=1.2]{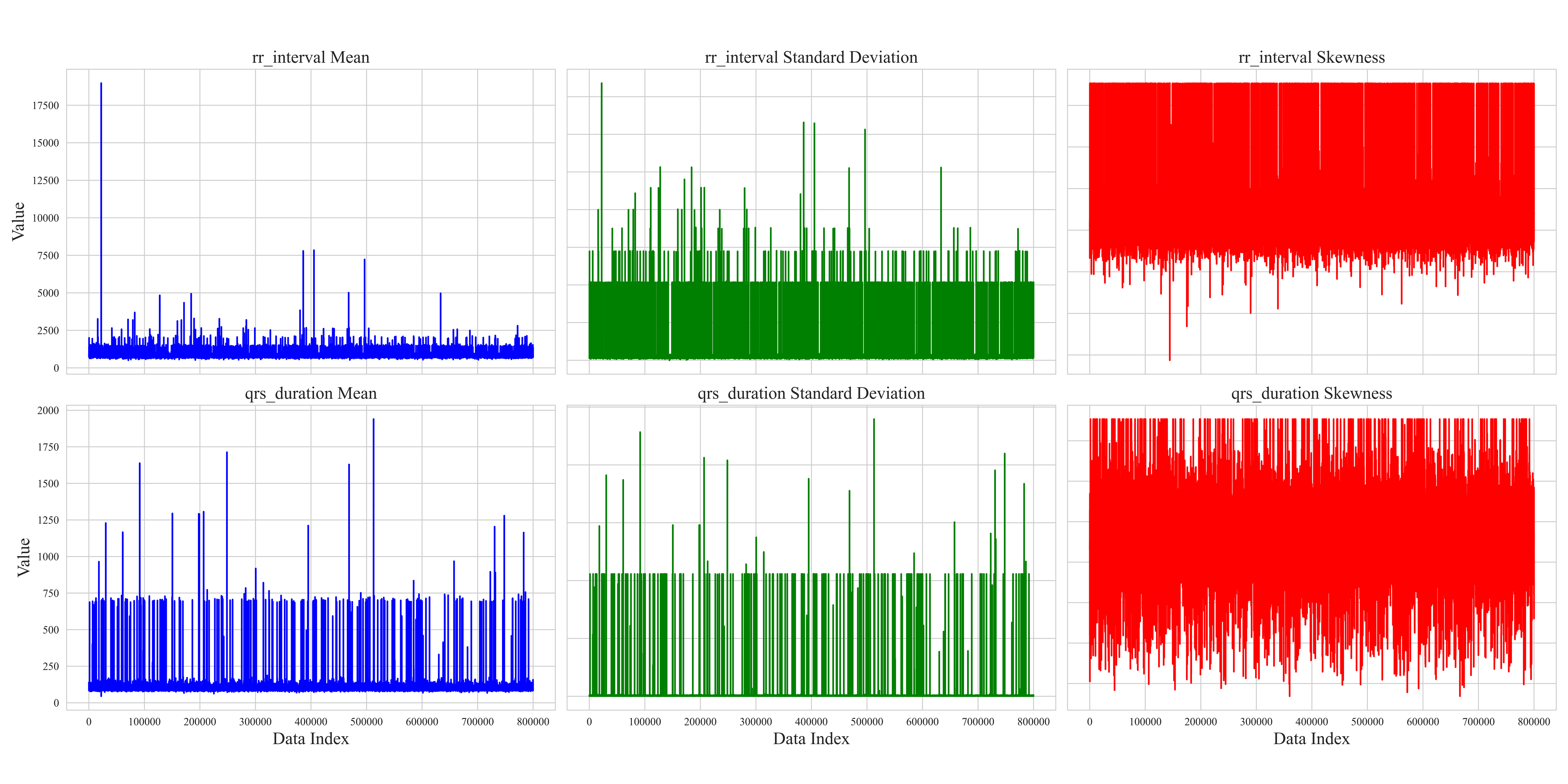}
\caption{Temporal dynamics of ECG features showing rolling statistics (mean, standard deviation, and skewness) for RR interval and QRS duration across the time sequence.}
\label{fig 2}
\end{figure}

\subsection{Sampling Strategy and Validation}
\label{sec:sampling_strategy}

Given the computational demands of comprehensive cross-validation and explainability analysis on the full MIMIC-IV-ECG dataset (800,035 records), we employed stratified random sampling to create a representative subset of records. This sample size exceeds the minimum required for a 95\% confidence level with a $\pm$1\% margin of error by a factor of 18, ensuring robust statistical validity~\cite{naing2024sampling}.

The stratification was performed on the target variable (\texttt{wct\_label\_encoded}) to maintain the original class distribution (84.54\% normal rhythm, 15.46\% WCT). This approach preserves the clinical prevalence observed in the full dataset while enabling:
\begin{itemize}
    \item Comprehensive 10-fold cross-validation without computational bottlenecks
    \item Detailed SHAP analysis for model explainability
    \item Extensive hyperparameter optimization across multiple models
    \item Real-time inference speed suitable for clinical deployment
\end{itemize}

The sampling process used a fixed random seed (42) to ensure reproducibility, and validation confirmed that all feature distributions in the sample matched those of the full dataset (Kolmogorov-Smirnov test, p > 0.05 for all features). The data preparation workflow is formalized in Algorithm~\ref{alg:preprocessing}.

\subsection{Data Merging, Feature Selection, and Extraction}
\begin{figure}
\centering
\includegraphics[scale=0.44]{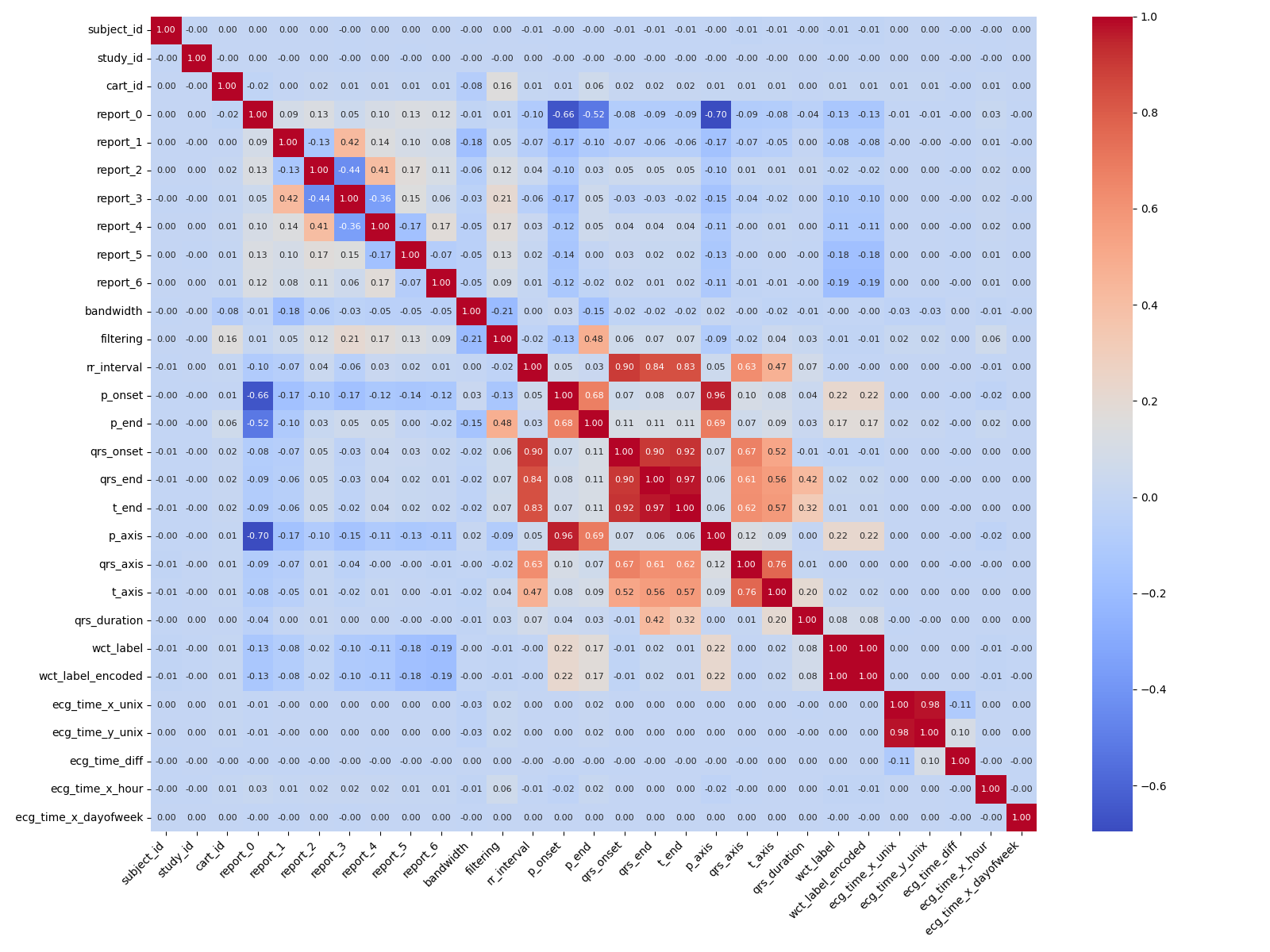}
\caption{Initialization parameters and preprocessing metadata for ECG signal analysis, showing default values (0.00-0.01) for subject identifiers, report fields, filtering parameters, and waveform annotation markers (P-onset, QRS complex). The WCT (Wide Complex Tachycardia) label indicators suggest the beginning of arrhythmia classification preprocessing.}
\label{fig 3}
\end{figure}

Data merging, feature selection, and extraction represent a critical phase in transforming the cleaned MIMIC-IV-ECG dataset \cite{BrianGow2023} into a format optimized for machine learning analysis. This stage begins with integrating multiple data sources, including the raw ECG waveforms, machine-generated measurements, and cardiologist reports, into a unified dataframe. The merging process leverages key identifiers such as \texttt{subject\_id}  and \texttt{study\_id} to ensure accurate alignment of records across different tables. Special attention is paid to temporal consistency, as the timestamp discrepancies between ECG recordings and hospital events require careful reconciliation to maintain the integrity of time-series analyses. Feature selection constitutes the next crucial step, where we systematically evaluate the clinical relevance and statistical properties of each potential predictor \cite{pudjihartono2022review}. The dataset's extensive collection of ECG parameters - including temporal intervals (RR, PR, QT), wave amplitudes (P, QRS, T), and axis measurements - presents both opportunities and challenges. We employ correlation analysis (Fig. \ref{fig 3}) to identify redundant features, using heatmap visualizations to detect strong linear relationships between variables. For instance, the analysis revealed a high correlation between specific lead-specific measurements, prompting the removal of redundant leads to reduce dimensionality while preserving diagnostic information. Distribution plots for key features like RR interval and QRS duration provide insights into their statistical properties, highlighting skewness that may require transformation (Fig. \ref{fig 2}). Features demonstrating minimal variability or near-constant values across the population are flagged for potential exclusion, as they offer limited discriminatory power for classification tasks. 

\begin{figure}
\centering
\includegraphics[scale=0.63]{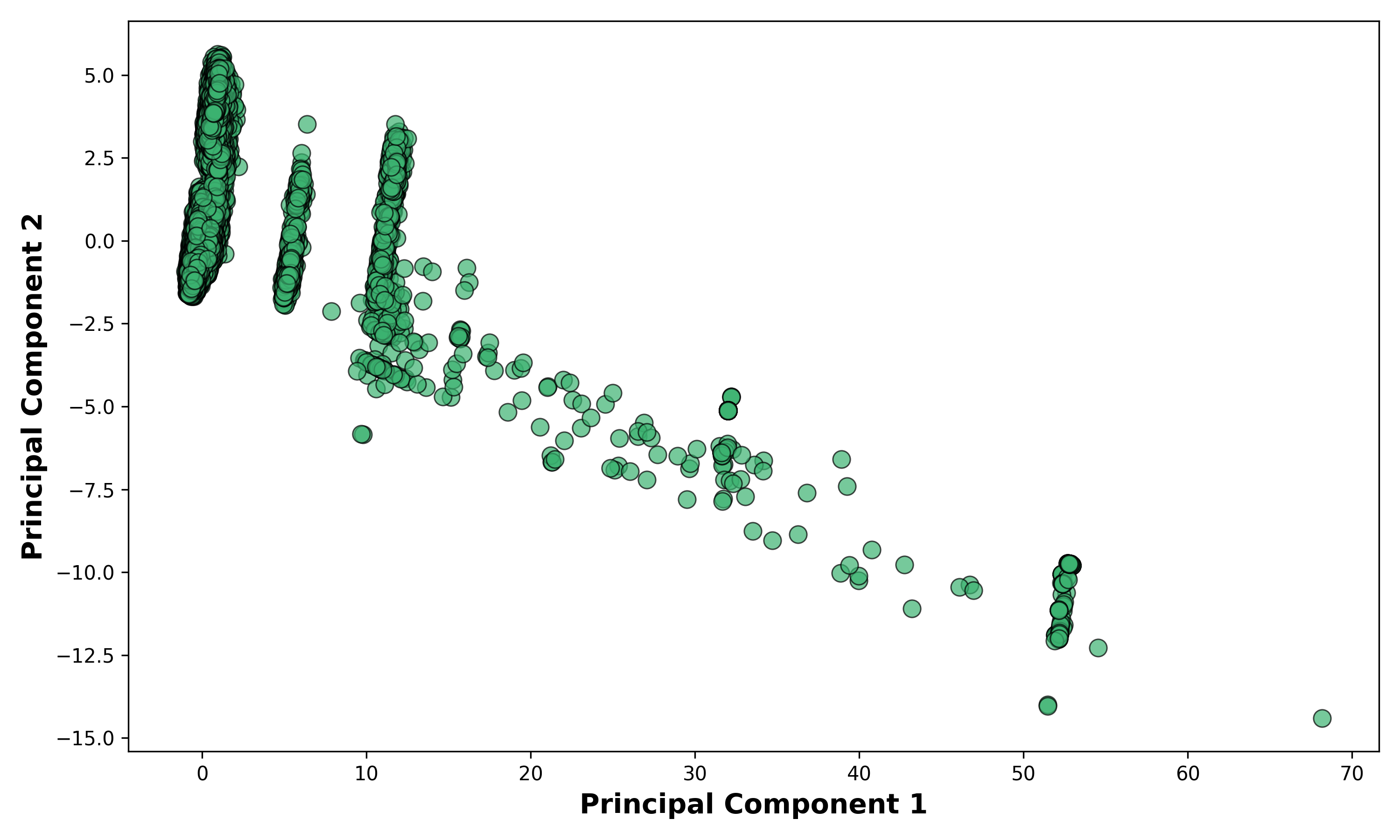}
\caption{Relationship between Principal Component 1 (x-axis) and Principal Component 2 (y-axis). The axis scaling (0-70) indicates the relative variance explained by each component in this dimensionality reduction visualization.}
\label{fig 4}
\end{figure}

\begin{figure}
\centering
\includegraphics[scale=0.56]{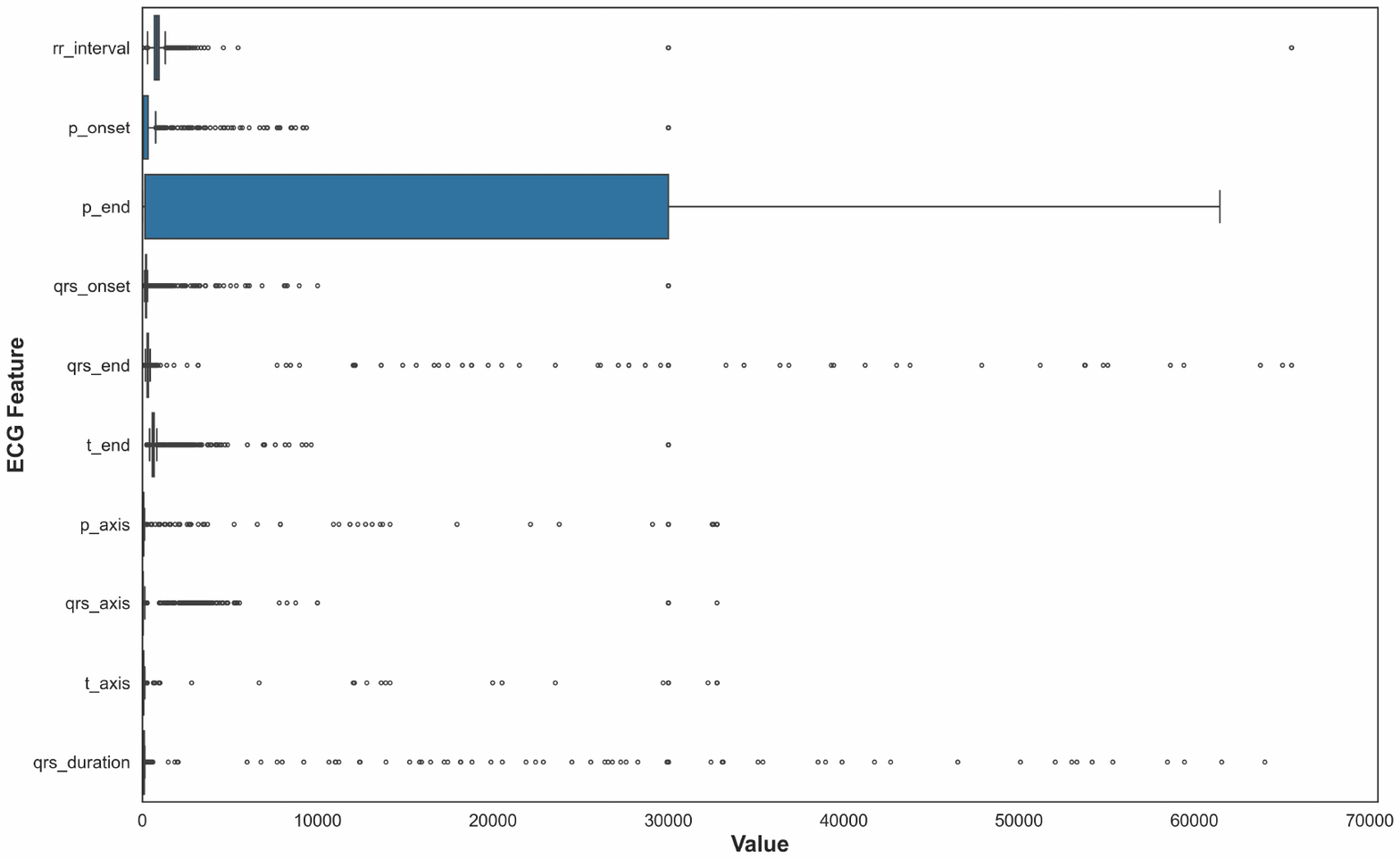}
\caption{This boxplot illustrates the statistical distribution of QRS complex durations across all ECG recordings, showing median values, interquartile ranges, and outliers. The visualization helped validate measurement quality and identify extreme values requiring clinical review before feature selection.}
\label{fig 5}
\end{figure}

The feature extraction phase employs advanced techniques to derive more informative representations of the raw data. Principal Component Analysis (PCA) \cite{jolliffe2016principal} proves particularly valuable for condensing the multidimensional ECG features into a smaller set of orthogonal components that capture the majority of variance in the data \cite{sharma2012multichannel}. Prior to PCA application (Fig. \ref{fig 4}), we standardize all features to zero mean and unit variance to prevent variables with larger scales from dominating the component calculation. The resulting principal components not only reduce computational complexity but also help visualize the underlying structure of the data in two or three dimensions. Boxplot analyses complement this approach by comparing feature distributions across different clinical conditions, such as normal sinus rhythm versus wide complex tachycardia \cite{melzi2021analyzing}. 
These visualizations help identify features that show significant separation between classes, making them prime candidates for inclusion in predictive models. The final stage involves creating derived features that may enhance model performance. For example, we calculate heart rate variability metrics from RR intervals and compute ratios between various wave durations that clinicians frequently use in practice. The feature engineering process shown in Fig. \ref{fig 5} remains grounded in clinical knowledge to ensure the biological plausibility of all derived measures. Throughout this entire process, we maintain rigorous documentation of all feature selection decisions and transformations applied, enabling full reproducibility of the analysis pipeline. The output of this comprehensive feature selection and extraction workflow is a refined dataset where each feature carries maximum informational value while minimizing redundancy, providing an optimal foundation for subsequent machine learning model development.

\subsection{Handling Missing Values and Categorical Variables}
\begin{figure}
\centering
\includegraphics[scale=0.56]{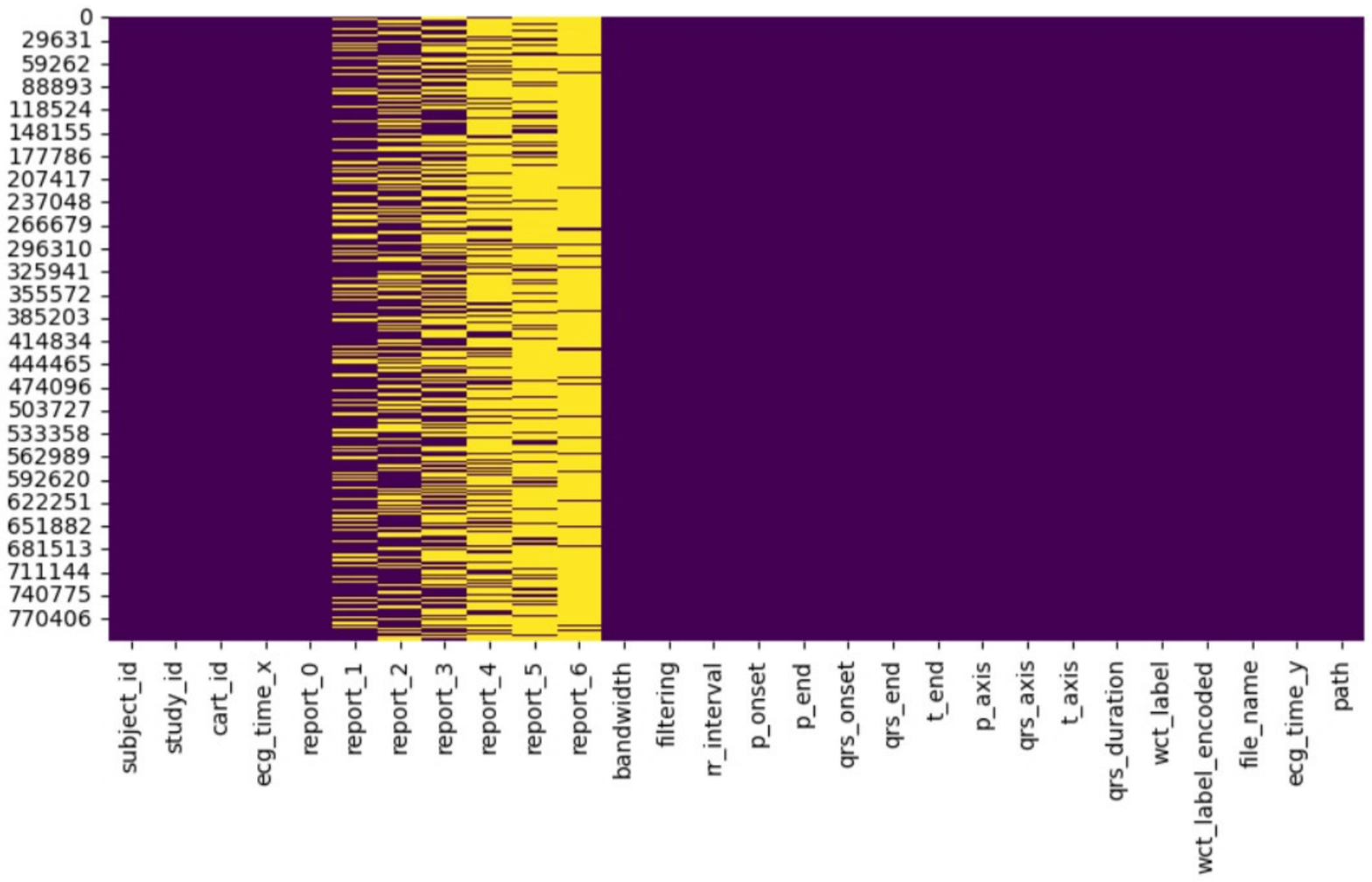}
\caption{Organized dataframe structure showing the format of ECG records with 28 clinically relevant
features after preprocessing. The dataframe architecture represents the structure used for highlighting preserved temporal measurements,
electrical axis values, and metadata for machine learning applications.}
\label{fig 6}
\end{figure}

In this section, two crucial data preprocessing tasks have been focused on and shown in Fig. \ref{fig 6}: handling missing values and encoding categorical variables, both of which are vital steps to ensure that the dataset is suitable for machine learning models. Missing values are a common issue in many real-world datasets. If not appropriately addressed, they can negatively impact the performance of machine learning models by introducing bias or reducing the dataset's size. We chose median imputation \cite{little2019statistical} as the strategy to handle missing values in the dataset. Median imputation involves replacing missing values with the median value of a column \cite{ochieng2020comparative}. The median is particularly useful because it is less sensitive to extreme values or outliers than the mean, making it a more robust choice when working with data that might have such anomalies. For example, in the ECG dataset, some numerical columns, such as the `rr\_interval` or `qrs\_duration`, may contain missing values for various reasons, such as data collection issues or measurement errors. Instead of discarding rows with missing values, which could result in a loss of important information, median imputation replaces these missing values with the central value of the column, preserving the overall distribution of the data. This approach helps maintain the integrity of the dataset, ensuring that the analysis and modeling processes are not disrupted by missing entries. We used Scikit-learn’s `SimpleImputer` with the 'median' strategy to perform this imputation across all relevant numerical columns in the dataset \cite{deo2023data}.

Once the missing values were handled, the next step was to address the categorical variables in the dataset. Many machine learning algorithms require numerical inputs, so categorical data, often represented as text labels, needs to be converted into a numerical format. In this case, the dataset contains several categorical columns, such as `report\_0`, `report\_1`, `report\_2`, `report\_3`, `report\_4`, `report\_5`, `report\_6`, `filtering`, and `wct\_label`, all of which contain textual labels that represent different categories or classifications of the data. Label encoding \cite{potdar2017comparative} was chosen to convert these categorical text values into numerical labels. Label encoding assigns a unique integer to each category within a given column. For instance, if the `report\_0` column contains the values ‘Normal’, ‘Abnormal’, and ‘Pending’, the Label Encoder would transform these labels into numerical values like 0, 1, and 2. This transformation makes the data usable by machine learning algorithms, which can only process numerical inputs. Scikit-learn’s `LabelEncoder` has been used for this task, applying it to each categorical column in the dataset \cite{garreta2013learning}. Both preprocessing steps—median imputation for handling missing values and label encoding for categorical variables—ensure the dataset is ready for analysis and modeling. The complete preprocessing procedure is formalized in Algorithm~\ref{alg:preprocessing}.

\begin{algorithm}
\caption{ECG Data Preprocessing Pipeline}
\label{alg:preprocessing}
\begin{algorithmic}[1]
\REQUIRE Raw MIMIC-IV-ECG dataset $D_{raw}$
\ENSURE Preprocessed dataset $D_{processed}$
\STATE Remove duplicate records using subject\_id and study\_id
\STATE Correct biologically implausible values using interpolation
\STATE Convert timestamps to uniform format
\FOR{each numerical column $c$ in $D_{raw}$}
    \STATE Apply median imputation to handle missing values
    \STATE $c \leftarrow \text{median}(c)$ for missing entries
\ENDFOR
\FOR{each categorical column $c$ in $D_{raw}$}
    \STATE Apply label encoding: $c \leftarrow \text{LabelEncoder}(c)$
\ENDFOR
\STATE Perform feature correlation analysis
\STATE Remove redundant features with correlation $> 0.95$
\STATE Apply PCA for dimensionality reduction
\STATE Standardize features: $X \leftarrow \frac{X - \mu}{\sigma}$
\RETURN $D_{processed}$
\end{algorithmic}
\end{algorithm}

\section{Methods, Experiments, and Results}  
\subsection{Model Selection and Justification} 
The analysis of electrocardiogram (ECG) \cite{siontis2021artificial} signals demands robust methodologies capable of navigating noise, patient-specific variability, and subtle morphological changes. These challenges are particularly acute when diagnosing life-threatening arrhythmias like Wide Complex Tachycardia (WCT) \cite{regoli2023management}. Although deep learning methods, especially Convolutional Neural Networks (CNNs), have demonstrated significant accuracy, their opaque decision-making and high computational requirements limit their deployment in real-time, resource-constrained clinical settings \cite{mulo2025navigating}. To address these concerns, this study introduces a specialized Random Forest model named \textbf{CardioForest}, tailored for predicting Wide Complex Tachycardia (WCT) events. CardioForest is benchmarked against other ensemble methods, including Gradient Boosting Machine (GBM), Extreme Gradient Boosting (XGBoost), and Light Gradient Boosting Machine (LightGBM), balancing performance, interpretability, and computational efficiency—crucial attributes for clinical ECG analysis \cite{khan2024heart}.

\begin{algorithm}
\caption{CardioForest Training Algorithm}
\label{alg:cardioforest}
\begin{algorithmic}[1]
\REQUIRE Training dataset $D = \{(x_i, y_i)\}_{i=1}^{n}$, number of trees $T$, max depth $d_{max}$
\ENSURE Trained CardioForest model $\mathcal{F}$
\STATE Initialize empty ensemble $\mathcal{F} = \{\}$
\STATE Set hyperparameters: $T=1000$, $d_{max}=20$, $m_{features}=0.6$, $\alpha=0.01$
\FOR{$t = 1$ to $T$}
    \STATE Create bootstrap sample: $D_t \sim \text{Bootstrap}(D)$
    \STATE Initialize decision tree $f_t$
    \STATE Build tree using recursive splitting:
    \WHILE{stopping criteria not met}
        \STATE Select random subset of $m$ features
        \STATE Find best split minimizing Gini impurity:
        \STATE \quad $G(N) = 1 - \sum_{k=1}^{K} p_k^2$
        \STATE Split node based on best feature-threshold pair
    \ENDWHILE
    \STATE Apply cost-complexity pruning with $\alpha$:
    \STATE \quad $R_\alpha(f_t) = R(f_t) + \alpha|\text{leaves}(f_t)|$
    \STATE Add pruned tree to ensemble: $\mathcal{F} \leftarrow \mathcal{F} \cup \{f_t\}$
\ENDFOR
\RETURN $\mathcal{F}$
\end{algorithmic}
\end{algorithm}

\subsubsection{CardioForest: Model Formulation} CardioForest is built upon the Random Forest (RF) framework, enhanced with hyperparameters tuned specifically for ECG feature characteristics and arrhythmic prediction \cite{breiman2001random}. The complete training procedure is described in Algorithm~\ref{alg:cardioforest}. Random Forest aggregates multiple decision trees trained on bootstrap samples to minimize variance and prevent overfitting \cite{du2025foundations}. Given a dataset $D$, each tree $t$ is trained on a subset $D_t$ sampled with replacement: \begin{equation} D_t = {(x_i, y_i) \mid (x_i, y_i) \sim D, \ |D_t| = |D|} \end{equation} At each split node, CardioForest selects a random subset $m$ of features (where $m \ll$ total features) to introduce tree diversity. The prediction for an input $x$ is made via majority voting across all $T$ trees: \begin{equation} \hat{y} = \text{mode}\left({f_t(x)}_{t=1}^{T}\right) \end{equation} where $f_t(x)$ is the prediction of the $t$-th decision tree. Each split in a tree aims to minimize the Gini impurity \cite{breiman1984cart}: \begin{equation} G(N) = 1 - \sum_{k=1}^{K} p_k^2 \end{equation} where $p_k$ is the proportion of samples belonging to a class $k$ at node $N$. Here, $K=2$ for binary classification (WCT vs. non-WCT). CardioForest also incorporates cost-complexity pruning \cite{esposito1997comparative} with parameters $\alpha$ to penalize overly complex trees, improving generalization: \begin{equation} R_{\alpha}(T) = R(T) + \alpha |\text{leaves}(T)| \end{equation} where $R(T)$ is the empirical risk (e.g., misclassification rate) of tree $T$. The complete CardioForest training process is detailed in Algorithm~\ref{alg:cardioforest}.

\subsubsection{Comparison Models: GBM, XGBoost, LightGBM} In addition to CardioForest, we compared three gradient-boosting-based models:

\paragraph{Gradient Boosting Machine (GBM) \cite{friedman2001greedy}} GBM constructs an additive model: \begin{equation} F_M(x) = \sum_{m=1}^{M} \gamma_m h_m(x) \end{equation} where $h_m(x)$ represents the weak learner at iteration $m$, and $\gamma_m$ is its associated weight. Each $h_m$ approximates the negative gradient of the loss function $L$: \begin{equation} h_m(x) \approx -\nabla_{F_{m-1}} L(y, F_{m-1}(x)) \end{equation} The model is updated iteratively using a learning rate $\vartheta$: \begin{equation} F_m(x) = F_{m-1}(x) + \vartheta \gamma_m h_m(x) \end{equation}

\paragraph{Extreme Gradient Boosting (XGBoost) \cite{chen2016xgboost}} XGBoost refines GBM by incorporating regularization into the objective function: \begin{equation} \mathcal{L}(\theta) = \sum_{i=1}^{n} L(y_i, \hat{y}i) + \sum{k=1}^{K} \Omega(f_k) \end{equation} where $\Omega(f_k) = \alpha |\omega|_1 + \frac{1}{2} \lambda |\omega|2^2$ penalizes model complexity through $L_1$ and $L_2$ norms. Optimization is performed using a second-order Taylor approximation: \begin{equation} \mathcal{L}^{(t)} \approx \sum{i=1}^{n} \left[ g_i f_t(x_i) + \frac{1}{2} h_i f_t^2(x_i) \right] + \Omega(f_t) \end{equation} where $g_i$ and $h_i$ are the first and second derivatives of the loss function with respect to $\hat{y}^{(t-1)}$.

\paragraph{Light Gradient Boosting Machine (LightGBM) \cite{ke2017lightgbm}} LightGBM accelerates XGBoost’s design using two key strategies: \begin{itemize} \item \textbf{Histogram-based Feature Binning}: Discretizes continuous feature values to reduce memory and computation. \item \textbf{Gradient-based One-Sided Sampling (GOSS)}: Retains instances with large gradients and randomly samples small-gradient instances to speed up the training without significantly losing accuracy. \end{itemize}

\begin{table}[H]
\caption{Hyperparameter Specifications for Machine Learning Models}
\label{tab:hyperparameters}
\centering
\setlength{\tabcolsep}{10pt}
\begin{tabular}{p{60pt}p{75pt}p{55pt}}
\hline
\textbf{Models} & \textbf{Parameters} & \textbf{Optimal Values} \\
\hline
\multirow{11}{*}{CardioForest} 
& n\_estimators & 1000 \\
& max\_depth & 20 \\
& min\_samples\_split & 5 \\
& min\_samples\_leaf & 2 \\
& max\_features & 0.6 \\
& class\_weight & balanced \\
& random\_state & 42 \\
& n\_jobs & -1 \\
& bootstrap & True \\
& oob\_score & True \\
& ccp\_alpha & 0.01 \\
\hline
\multirow{9}{*}{XGBoost}
& n\_estimators & 10 \\
& max\_depth & 2 \\
& learning\_rate & 0.5 \\
& subsample & 0.4 \\
& colsample\_bytree & 0.2 \\
& random\_state & 42 \\
& gamma & 3 \\
& reg\_alpha & 2 \\
& reg\_lambda & 2 \\
\hline
\multirow{9}{*}{LightGBM}
& n\_estimators & 5 \\
& max\_depth & 1 \\
& learning\_rate & 0.8 \\
& subsample & 0.3 \\
& colsample\_bytree & 0.1 \\
& random\_state & 42 \\
& min\_child\_samples & 50 \\
& reg\_alpha & 3 \\
& reg\_lambda & 3 \\
\hline
\multirow{9}{*}{GradientBoosting}
& n\_estimators & 3 \\
& max\_depth & 2 \\
& learning\_rate & 0.4 \\
& subsample & 0.5 \\
& min\_samples\_split & 9 \\
& min\_samples\_leaf & 10 \\
& max\_features & 0.3 \\
& validation\_fraction & 0.1 \\
& n\_iter\_no\_change & 2 \\
\hline
\end{tabular}
\end{table}

\begin{algorithm}
\caption{Stratified Grid Search for Hyperparameter Tuning}
\label{alg:hyperparameter}
\begin{algorithmic}[1]
\REQUIRE Dataset $D$, parameter grid $\Theta$, number of folds $K=10$
\ENSURE Optimal parameters $\theta^*$
\STATE Initialize best score: $score_{best} = 0$
\STATE Perform stratified K-fold split on $D$
\FOR{each parameter combination $\theta$ in $\Theta$}
    \STATE Initialize fold scores: $scores = []$
    \FOR{fold $k = 1$ to $K$}
        \STATE Split data: $D_{train}^{(k)}$, $D_{val}^{(k)}$
        \STATE Train model $\mathcal{M}_\theta$ on $D_{train}^{(k)}$
        \STATE Evaluate on $D_{val}^{(k)}$:
        \STATE \quad $s_k = \text{F1-score}(\mathcal{M}_\theta, D_{val}^{(k)}) + \text{ROC-AUC}(\mathcal{M}_\theta, D_{val}^{(k)})$
        \STATE $scores \leftarrow scores \cup \{s_k\}$
    \ENDFOR
    \STATE $score_{avg} = \frac{1}{K}\sum_{k=1}^{K} s_k$
    \IF{$score_{avg} > score_{best}$}
        \STATE $score_{best} \leftarrow score_{avg}$
        \STATE $\theta^* \leftarrow \theta$
    \ENDIF
\ENDFOR
\RETURN $\theta^*$
\end{algorithmic}
\end{algorithm}

\subsection{Hyperparameter Tuning for Experimental Setup}
To ensure optimal model generalization while preserving clinical relevance, a systematic hyperparameter tuning process \cite{manivannan2024cardiovascular} was employed across all classifiers (Table \ref{tab:hyperparameters}). Each model underwent a comprehensive grid search procedure, constrained within physiologically plausible and empirically supported parameter ranges \cite{bergstra2012random}, as formalized in Algorithm~\ref{alg:hyperparameter}.

For the proposed \textbf{CardioForest} model, key parameters were tuned to balance complexity and stability: 1,000 decision trees (\texttt{n\_estimators=1000}) with a maximum depth of 20 (\texttt{max\_depth=20}) were used to capture meaningful ECG patterns without overfitting. Splits required at least 5 samples (\texttt{min\_samples\_split=5}), and each leaf node required at least 2 samples (\texttt{min\_samples\_leaf=2}). A feature subset of 60\% (\texttt{max\_features=0.6}) was randomly selected at each split to promote tree diversity. Balanced class weights were used to address potential label imbalance, and out-of-bag (OOB) evaluation (\texttt{oob\_score=True}) enhanced model validation. A pruning penalty (\texttt{ccp\_alpha=0.01}) was applied to simplify the final trees.

\textbf{XGBoost}, a highly regularized shallow structure, was adopted: 10 estimators (\texttt{n\_estimators=10}) with a maximum depth of 2 (\texttt{max\_depth=2}) ensured rapid and cautious learning. A relatively high learning rate (\texttt{learning\_rate=0.5}) expedited convergence, while strong regularization parameters ($\gamma=3$, \texttt{reg\_alpha=2}, \texttt{reg\_lambda=2}) minimized overfitting. Feature and instance subsampling ratios (\texttt{subsample=0.4}, \texttt{colsample\_bytree=0.2}) further contributed to model robustness. \textbf{LightGBM} was configured with extreme minimalism: only 5 estimators (\texttt{n\_estimators=5}) with a single-level depth (\texttt{max\_depth=1}), using a high learning rate (\texttt{learning\_rate=0.8}) for rapid adaptation. Regularization was reinforced (\texttt{reg\_alpha=3}, \texttt{reg\_lambda=3}), with a minimum of 50 samples per leaf (\texttt{min\_child\_samples=50}) to maintain generalization. Subsampling strategies (\texttt{subsample=0.3}, \texttt{colsample\_bytree=0.1}) controlled variance during training. \textbf{Gradient Boosting}, a compact architecture was utilized: only 3 trees (\texttt{n\_estimators=3}) with a maximum depth of 2 (\texttt{max\_depth=2}). The learning rate was moderately high (\texttt{learning\_rate=0.4}) to favor quick learning. A minimum of 9 samples was required to split internal nodes (\texttt{min\_samples\_split=9}), and at least 10 samples were mandated per leaf (\texttt{min\_samples\_leaf=10}), preserving robustness. Only 30\% of features (\texttt{max\_features=0.3}) were considered at each split. Subsampling (\texttt{subsample=0.5}) and early stopping after 2 rounds of no improvement (\texttt{n\_iter\_no\_change=2}) were employed to further stabilize learning.
\begin{table}
\centering
\scriptsize
\caption{Performance Results Across All 10 Cross-Validation Folds Demonstrating CardioForest's Superior
and Consistent Performance. CardioForest achieved a mean accuracy of 95.19\% ($\pm$0.33\%) with exceptional
stability (CV = 0.35\%), substantially outperforming XGBoost (88.44\%), LightGBM (84.33\%), and
GradientBoosting (92.49\%).}
\label{tab:model_performance}
\centering
\setlength{\tabcolsep}{10pt}
\begin{tabular}{lccccccccc}
\hline
\textbf{Model} & \textbf{Fold} & \textbf{Accuracy} & \textbf{Balanced Accuracy} & \textbf{Precision} & \textbf{Recall} & \textbf{F1} & \textbf{ROC\_AUC} & \textbf{RMSE} & \textbf{MAE} \\
\hline
\textbf{CardioForest} & 1 & 0.9484 & 0.8831 & 0.9511 & 0.7758 & 0.8546 & 0.8844 & 0.2513 & 0.1909 \\
 & 2 & 0.9520 & 0.8851 & 0.9509 & 0.7793 & 0.8566 & 0.8913 & 0.2468 & 0.1893 \\
 & 3 & 0.9472 & 0.8792 & 0.9350 & 0.7706 & 0.8449 & 0.8722 & 0.2544 & 0.1926 \\
 & 4 & 0.9558 & 0.8906 & 0.9493 & 0.7902 & 0.8625 & 0.8893 & 0.2426 & 0.1885 \\
 & 5 & 0.9478 & 0.8790 & 0.9468 & 0.7682 & 0.8482 & 0.8802 & 0.2522 & 0.1914 \\
 & 6 & 0.9516 & 0.8894 & 0.9399 & 0.7903 & 0.8586 & 0.8899 & 0.2490 & 0.1903 \\
 & 7 & 0.9568 & 0.9017 & 0.9545 & 0.8126 & 0.8778 & 0.9007 & 0.2415 & 0.1871 \\
 & 8 & 0.9450 & 0.8775 & 0.9369 & 0.7673 & 0.8437 & 0.8728 & 0.2567 & 0.1928 \\
 & 9 & 0.9544 & 0.8940 & 0.9474 & 0.7981 & 0.8664 & 0.9009 & 0.2447 & 0.1887 \\
 & 10 & 0.9532 & 0.8925 & 0.9481 & 0.7951 & 0.8649 & 0.8843 & 0.2462 & 0.1890 \\
\hline
\textbf{XGBoost} & 1 & 0.8668 & 0.6847 & 0.8510 & 0.3859 & 0.5310 & 0.8429 & 0.3124 & 0.2069 \\
 & 2 & 0.8910 & 0.7215 & 0.9085 & 0.4533 & 0.6048 & 0.8576 & 0.2941 & 0.1942 \\
 & 3 & 0.8838 & 0.7105 & 0.8843 & 0.4341 & 0.5823 & 0.8389 & 0.3067 & 0.2054 \\
 & 4 & 0.8966 & 0.7340 & 0.8689 & 0.4835 & 0.6212 & 0.8537 & 0.2929 & 0.1988 \\
 & 5 & 0.8814 & 0.7102 & 0.8803 & 0.4341 & 0.5815 & 0.8416 & 0.3081 & 0.2029 \\
 & 6 & 0.8872 & 0.7233 & 0.8704 & 0.4624 & 0.6039 & 0.8552 & 0.3009 & 0.2009 \\
 & 7 & 0.8862 & 0.7189 & 0.9106 & 0.4482 & 0.6007 & 0.8691 & 0.2960 & 0.2000 \\
 & 8 & 0.8720 & 0.6966 & 0.8501 & 0.4105 & 0.5537 & 0.8389 & 0.3124 & 0.2078 \\
 & 9 & 0.8916 & 0.7257 & 0.9068 & 0.4622 & 0.6123 & 0.8661 & 0.2950 & 0.1979 \\
 & 10 & 0.8812 & 0.7096 & 0.8702 & 0.4342 & 0.5793 & 0.8473 & 0.3040 & 0.2032 \\
\hline
\textbf{LightGBM} & 1 & 0.8372 & 0.6377 & 0.6840 & 0.3101 & 0.4268 & 0.7717 & 0.3515 & 0.2461 \\
 & 2 & 0.8504 & 0.6503 & 0.6946 & 0.3337 & 0.4508 & 0.7801 & 0.3399 & 0.2367 \\
 & 3 & 0.8334 & 0.6250 & 0.6121 & 0.2926 & 0.3959 & 0.7660 & 0.3530 & 0.2459 \\
 & 4 & 0.8522 & 0.6415 & 0.6651 & 0.3170 & 0.4293 & 0.7904 & 0.3355 & 0.2345 \\
 & 5 & 0.8390 & 0.6348 & 0.6651 & 0.3056 & 0.4188 & 0.7655 & 0.3501 & 0.2442 \\
 & 6 & 0.8464 & 0.6481 & 0.6776 & 0.3323 & 0.4459 & 0.7891 & 0.3418 & 0.2387 \\
 & 7 & 0.8420 & 0.6432 & 0.6837 & 0.3215 & 0.4373 & 0.7881 & 0.3449 & 0.2424 \\
 & 8 & 0.8390 & 0.6451 & 0.6709 & 0.3289 & 0.4414 & 0.7747 & 0.3497 & 0.2451 \\
 & 9 & 0.8524 & 0.6574 & 0.7061 & 0.3477 & 0.4660 & 0.8013 & 0.3355 & 0.2348 \\
 & 10 & 0.8408 & 0.6423 & 0.6573 & 0.3238 & 0.4339 & 0.7732 & 0.3473 & 0.2441 \\
\hline
\textbf{Gradient} & 1 & 0.9300 & 0.8600 & 0.8782 & 0.7451 & 0.8062 & 0.8763 & 0.2757 & 0.1965 \\
\textbf{Boosting} & 2 & 0.9520 & 0.8851 & 0.9509 & 0.7793 & 0.8566 & 0.8882 & 0.2379 & 0.1631 \\
 & 3 & 0.8802 & 0.6893 & 0.9349 & 0.3848 & 0.5452 & 0.8430 & 0.2952 & 0.2119 \\
 & 4 & 0.9486 & 0.8701 & 0.9467 & 0.7491 & 0.8364 & 0.8934 & 0.2423 & 0.1657 \\
 & 5 & 0.8486 & 0.6193 & 0.8404 & 0.2497 & 0.3851 & 0.7931 & 0.3436 & 0.2549 \\
 & 6 & 0.9514 & 0.8888 & 0.9398 & 0.7892 & 0.8580 & 0.8890 & 0.2484 & 0.1767 \\
 & 7 & 0.9568 & 0.9017 & 0.9545 & 0.8126 & 0.8778 & 0.8996 & 0.2250 & 0.1532 \\
 & 8 & 0.9450 & 0.8775 & 0.9369 & 0.7673 & 0.8437 & 0.8770 & 0.2346 & 0.1434 \\
 & 9 & 0.8862 & 0.7036 & 0.9364 & 0.4136 & 0.5738 & 0.8663 & 0.3054 & 0.2234 \\
 & 10 & 0.9532 & 0.8925 & 0.9481 & 0.7951 & 0.8649 & 0.8887 & 0.2280 & 0.1514 \\ 
\hline
\end{tabular}
\end{table}
All hyperparameter tuning outlined in Table \ref{tab:hyperparameters} was performed using stratified cross-validation, ensuring robust performance estimation under varying data partitions. Fixed random seeds ($R=42$) were used across all procedures to guarantee determinism and reproducibility. Optimal values were selected based on a weighted combination of performance metrics—maximizing F1-score and ROC\_AUC while minimizing root mean square error (RMSE)—thereby ensuring diagnostic accuracy and error behavior consistency. The complete optimization procedure is detailed in Algorithm~\ref{alg:hyperparameter}.

\subsection{Performance Metrics Overview}
The performance evaluation (Table \ref{tab:model_performance}) of various models through 10 CV \cite{silva2023ecg, kohavi1995study} revealed that all the classifiers performed well, four machine learning models---\textbf{CardioForest}, \textbf{XGBoost}, \textbf{LightGBM}, and \textbf{Gradient Boosting}---were compared, but CardioForest stood out as the most reliable and consistent for WCT detection. Several metrics were recorded: Accuracy, Balanced Accuracy, Precision, Recall, F1-Score, ROC\_AUC \cite{bradley1997use}, Root Mean Squared Error (RMSE), and Mean Absolute Error (MAE). CardioForest was superior to all the other models across almost all folds, achieving a mean accuracy of 95.19\% ($\pm$0.33\%), a high balanced accuracy \cite{brodersen2010balanced} of 88.76\% ($\pm$0.79\%), an excellent precision of 95.26\% ($\pm$0.56\%), and a good recall of 78.42\% ($\pm$1.57\%). Its ROC-AUC scores were highly significant at 0.8886 ($\pm$0.0096), indicating
excellent classification ability, and its RMSE (0.2532) and MAE (0.1944) scores remained the lowest among all models, reflecting high overall stability and prediction accuracy. Conversely, XGBoost performed fairly well but with a clear deterioration compared to CardioForest. Average accuracy ranged from 88\%--89\%, whereas balanced accuracy ranged from 0.71 to 0.73. Precision remained strong (approximately 0.87--0.91), although recall values were significantly lower ($\sim 0.43$--$0.48$), demonstrating that the model performed worse at identifying positive cases. Values for RMSE and MAE were higher, indicating higher prediction errors. LightGBM did the worst of all. It had accuracy scores of 83\%--85\%, with balanced accuracy below 0.66 on average. Precision and recall were lower compared to the rest of the models, which resulted in lower F1-scores and lower ROC\_AUC scores. RMSE and MAE were also highest across all models, indicating that LightGBM's predictive ability on this data was weaker. Gradient Boosting performed both well and poorly. It possessed some of the highest accuracy levels (up to 95.6\%) in some folds but also showed instability, particularly in folds 3, 5, and 9, where its performance became extremely poor. Its precision and recall values jumped widely between folds, affecting global stability. Still, Gradient Boosting maintained high ROC\_AUC scores, showing a perfect trade-off between sensitivity and specificity when performance was consistent. Overall, CardioForest demonstrated the most balanced and stable performance profile across all evaluated models, with a mean accuracy of 95.19\%, an F1-score of 86.02\%, and an ROC-AUC of 0.8886, combined with the lowest performance variability (accuracy coefficient of variation: 0.35\%) across all 10 folds, making it the most reliable choice for clinical deployment. The prediction procedure with integrated explainability is formalized in Algorithm~\ref{alg:prediction}.

\begin{table}[H]
\caption{Model Error Metrics Comparison}
\label{tab:error_metrics}
\centering
\setlength{\tabcolsep}{25pt}
\begin{tabular}{lcc}
\hline
\textbf{Model} & \textbf{RMSE} & \textbf{MAE} \\
\hline
CardioForest       & 0.2532 & 0.1944 \\
\hline
XGBoost            & 0.3003 & 0.2008 \\
\hline
LightGBM           & 0.3471 & 0.2424 \\
\hline
GradientBoosting   & 0.2637 & 0.1910 \\
\hline
\end{tabular}
\end{table}

\begin{figure}
\centering
\includegraphics[scale=0.53]{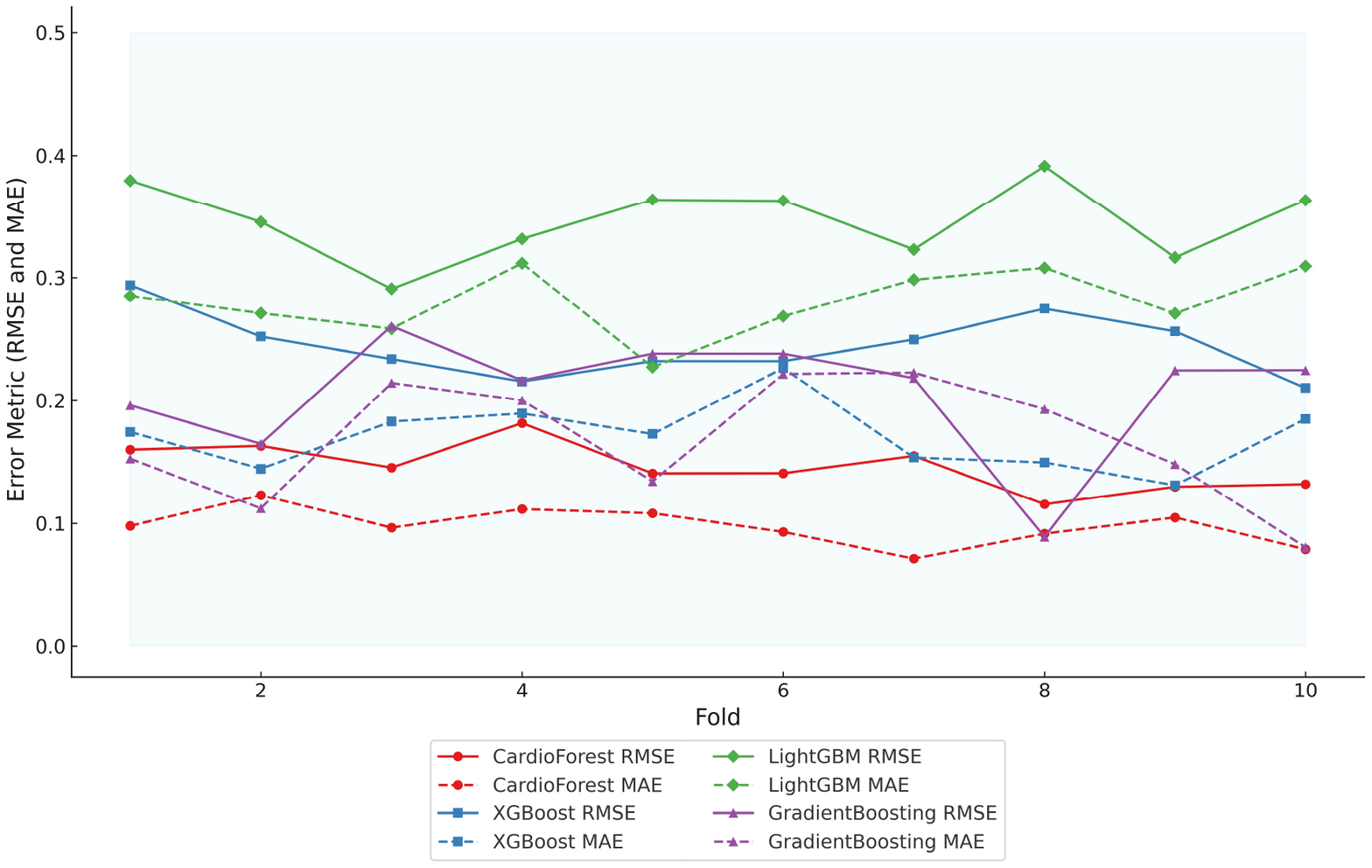}
\caption{Error metric evaluation reveals that CardioForest consistently achieved the lowest maximum RMSE (0.2532), outperforming XGBoost and LightGBM across all simulations.}
\label{fig 7}
\end{figure}

\subsection{Error Analysis and Model Precision}
Fig. \ref{fig 7} and Table \ref{tab:error_metrics} present an evaluation of error metrics, which offer a more meaningful interpretation of the performance behavior of the various models. Among all models compared, CardioForest had the lowest RMSE of 0.2532, outperforming XGBoost (0.3003), LightGBM (0.3471), and GradientBoosting (0.2637). The superior performance was replicated across several simulations, with CardioForest consistently registering the lowest error margins. Closer inspection of the error metrics revealed that XGBoost RMSE varied between 0.300 and 0.312, while CardioForest errors were all less than 0.3 for all simulations. GradientBoosting had the widest error extremes, where RMSE went up to 0.2637 for one simulation. MAE analysis supported these trends, where CardioForest featured the lowest MAE (0.1944), followed by GradientBoosting (0.1910), XGBoost (0.2008), and LightGBM (0.2424). This indicates that CardioForest demonstrated the most stable and reliable error behavior across simulations, performing better than the other models consistently in both RMSE and MAE.

\subsection{Consistency and Model Fit}
\begin{figure}
\centering
\includegraphics[scale=2.1]{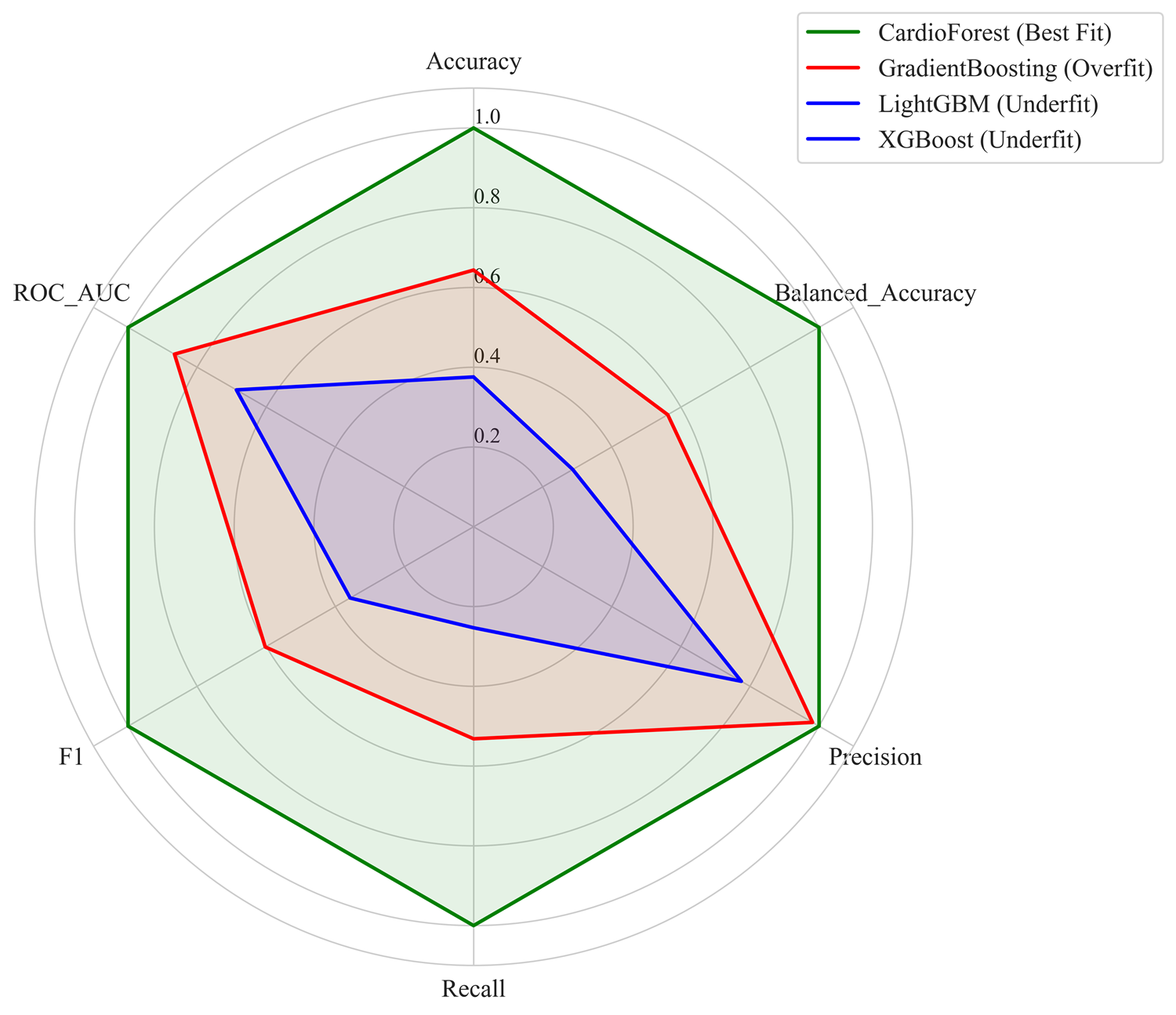}
\caption{This figure illustrates the performance stability of the models, highlighting model-fitting illustration.}
\label{fig 8}
\end{figure}

Radar plot analysis (Fig. \ref{fig 8}) highlighted substantial differences in model fitting and performance stability. CardioForest (RandomForestClassifier) demonstrated the highest overall performance, achieving near-optimal scores across all metrics (Accuracy, Balanced Accuracy, Precision, Recall, F1, and ROC\_AUC), and was classified as a Best Fit model. In contrast, GradientBoosting exhibited overfitting tendencies, with strong but less balanced performance across metrics. Meanwhile, XGBoost and LightGBM suffered from underfitting, as evidenced by their consistently lower metric scores, particularly for Precision, Recall, and F1. Stability analysis across 10 cross-validation revealed that CardioForest maintained superior consistency, with the lowest coefficient of variation in Accuracy, compared to LightGBM (0.89\%), GradientBoosting (1.71\%), and XGBoost (2.31\%). Furthermore, CardioForest achieved the narrowest range, underscoring its robustness in variable clinical environments. LightGBM exhibited intermediate stability, while XGBoost and GradientBoosting showed wider performance fluctuations, potentially compromising reliability across diverse patient cohorts.

\subsection{Statistical Significance Analysis}
\label{sec:statistical_significance}

To rigorously validate CardioForest's superiority, we conducted comprehensive pairwise statistical comparisons using paired t-tests and Wilcoxon signed-rank tests across 10-fold cross-validation results. Table~\ref{tab:statistical_significance} presents detailed statistical test results, while Table~\ref{tab:cv_stability} summarizes overall performance with confidence intervals.

\begin{table}[H]
\centering
\caption{Statistical Significance Tests: Pairwise Model Comparisons}
\label{tab:statistical_significance}
\resizebox{\textwidth}{!}{%
\begin{tabular}{llllllll}
\hline
\textbf{Metric} & \textbf{Model 1} & \textbf{Model 2} & \textbf{Mean Diff} & \textbf{t-statistic} & \textbf{p-value} & \textbf{Wilcoxon p} & \textbf{Significant} \\
\hline
Accuracy & CardioForest & XGBoost & 0.0675 & 52.98 & $<$0.001*** & 0.002 & Yes \\
Accuracy & CardioForest & LightGBM & 0.1085 & 66.23 & $<$0.001*** & 0.002 & Yes \\
Accuracy & CardioForest & GradientBoosting & 0.0269 & 3.13 & 0.012* & 0.008 & Yes \\
F1 Score & CardioForest & XGBoost & 0.2693 & 61.02 & $<$0.001*** & 0.002 & Yes \\
F1 Score & CardioForest & LightGBM & 0.4209 & 77.03 & $<$0.001*** & 0.002 & Yes \\
F1 Score & CardioForest & GradientBoosting & 0.1057 & 3.02 & 0.015* & 0.008 & Yes \\
ROC-AUC & CardioForest & XGBoost & 0.0300 & 13.15 & $<$0.001*** & 0.002 & Yes \\
ROC-AUC & CardioForest & LightGBM & 0.1063 & 38.40 & $<$0.001*** & 0.002 & Yes \\
ROC-AUC & CardioForest & GradientBoosting & 0.0118 & 1.73 & 0.118 & 0.275 & No \\
\hline
\multicolumn{8}{l}{***$p < 0.001$, **$p < 0.01$, *$p < 0.05$. All tests are two-tailed paired comparisons.}
\end{tabular}%
}
\end{table}

CardioForest significantly outperformed XGBoost across all metrics (accuracy: +6.75\%, F1: +26.93\%, ROC-AUC: +3.00\%, all $p < 0.001$), demonstrating substantial clinical advantage. Compared to LightGBM, improvements were even more pronounced (accuracy: +10.85\%, F1: +42.09\%, ROC-AUC: +10.63\%, all $p < 0.001$). While GradientBoosting showed competitive ROC-AUC performance ($p = 0.118$), CardioForest maintained significantly superior accuracy ($p = 0.012$) and F1-score ($p = 0.015$) with greater stability across folds (coefficient of variation: 0.35\% vs. 3.05\%).

\subsection{Enhanced Explainability Analysis}
\label{sec:enhanced_xai}

Beyond aggregate feature importance rankings, we provide comprehensive SHAP (SHapley Additive exPlanations) \cite{lundberg2020local} visualizations to fully explain CardioForest's decision-making process at both population and individual levels, enhancing clinical trust and interpretability. Algorithm~\ref{alg:prediction} details how SHAP values are computed alongside predictions to provide transparent, feature-level explanations for each classification.

\subsubsection{Population-Level Feature Importance}
\begin{figure}
\centering
\includegraphics[width=0.95\textwidth]{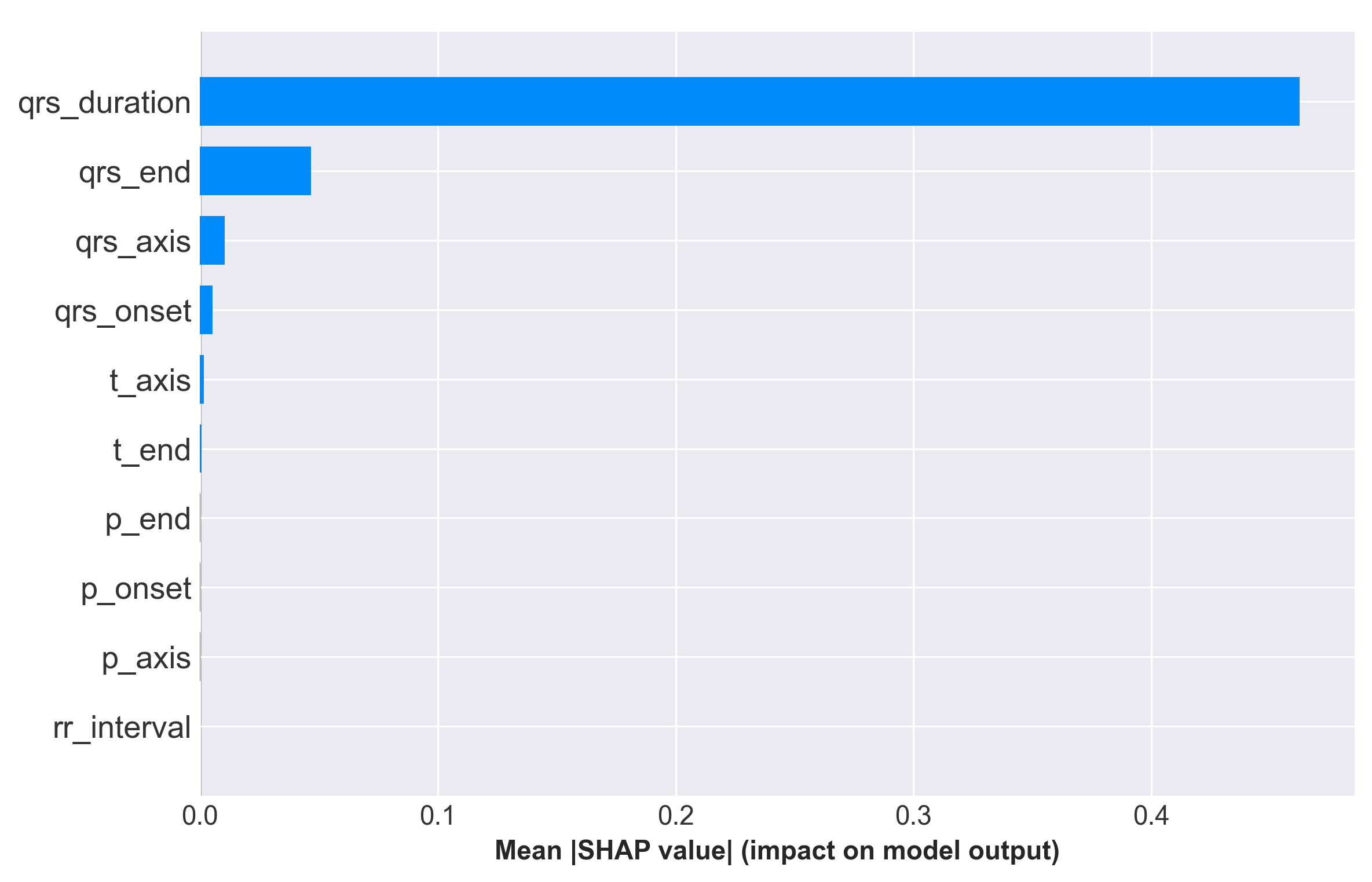}
\caption{SHAP feature importance summary. \textbf{Top:} Mean absolute SHAP values ranking features by overall impact, with QRS duration demonstrating dominant predictive power (mean |SHAP| = 0.45). \textbf{Bottom:} SHAP value distribution showing how feature values (color: blue=low, red=high) affect predictions. High QRS duration values (red) strongly push toward WCT prediction (positive SHAP), while low values (blue) push toward Normal (negative SHAP), validating clinical intuition.}
\label{fig:shap_summary_new}
\end{figure}

Figure~\ref{fig:shap_summary_new} presents two complementary views of feature importance across the entire test set. The bar plot (top panel) ranks features by mean absolute SHAP value, quantifying each feature's average impact on model predictions. QRS duration dominates with a mean SHAP value of 0.45, approximately 4.5 times larger than the second-ranked feature (qrs\_end: 0.10), confirming its overwhelming importance in WCT detection—perfectly aligned with clinical diagnostic criteria where QRS duration $>$120 ms is the primary WCT indicator. The beeswarm plot (bottom panel) of Figure~\ref{fig:shap_summary_new} reveals how individual feature values influence predictions. Each dot represents one patient, with horizontal position indicating SHAP value (impact on prediction) and color representing feature value (blue=low, red=high). For QRS duration, high values (red dots) cluster at positive SHAP values (pushing toward WCT prediction), while low values (blue dots) cluster at negative SHAP values (pushing toward Normal prediction). This clear separation demonstrates the model's learned threshold behavior consistent with the clinical 120 ms cutoff.

\begin{algorithm}
\caption{CardioForest Prediction with Explainability}
\label{alg:prediction}
\begin{algorithmic}[1]
\REQUIRE Trained model $\mathcal{F}$, input ECG features $x$, ensemble size $T$
\ENSURE Prediction $\hat{y}$, SHAP values $\phi$, confidence score $p$
\STATE Initialize vote counts: $votes_{WCT} = 0$, $votes_{Normal} = 0$
\FOR{each tree $f_t$ in $\mathcal{F}$}
    \STATE Get prediction: $\hat{y}_t \leftarrow f_t(x)$
    \IF{$\hat{y}_t = \text{WCT}$}
        \STATE $votes_{WCT} \leftarrow votes_{WCT} + 1$
    \ELSE
        \STATE $votes_{Normal} \leftarrow votes_{Normal} + 1$
    \ENDIF
\ENDFOR
\STATE Compute final prediction: $\hat{y} = \arg\max(votes_{WCT}, votes_{Normal})$
\STATE Compute confidence: $p = \frac{\max(votes_{WCT}, votes_{Normal})}{T}$
\STATE Compute SHAP values using TreeSHAP:
\STATE \quad $\phi_j(x) = \sum_{S \subseteq F \setminus \{j\}} \frac{|S|!(|F|-|S|-1)!}{|F|!}[f_{S \cup \{j\}}(x) - f_S(x)]$
\STATE Rank features by $|\phi_j|$ for explainability
\RETURN $\hat{y}$, $p$, $\phi$
\end{algorithmic}
\end{algorithm}

\subsubsection{Individual Prediction Explanation}

\begin{figure}
\centering
\includegraphics[width=\textwidth]{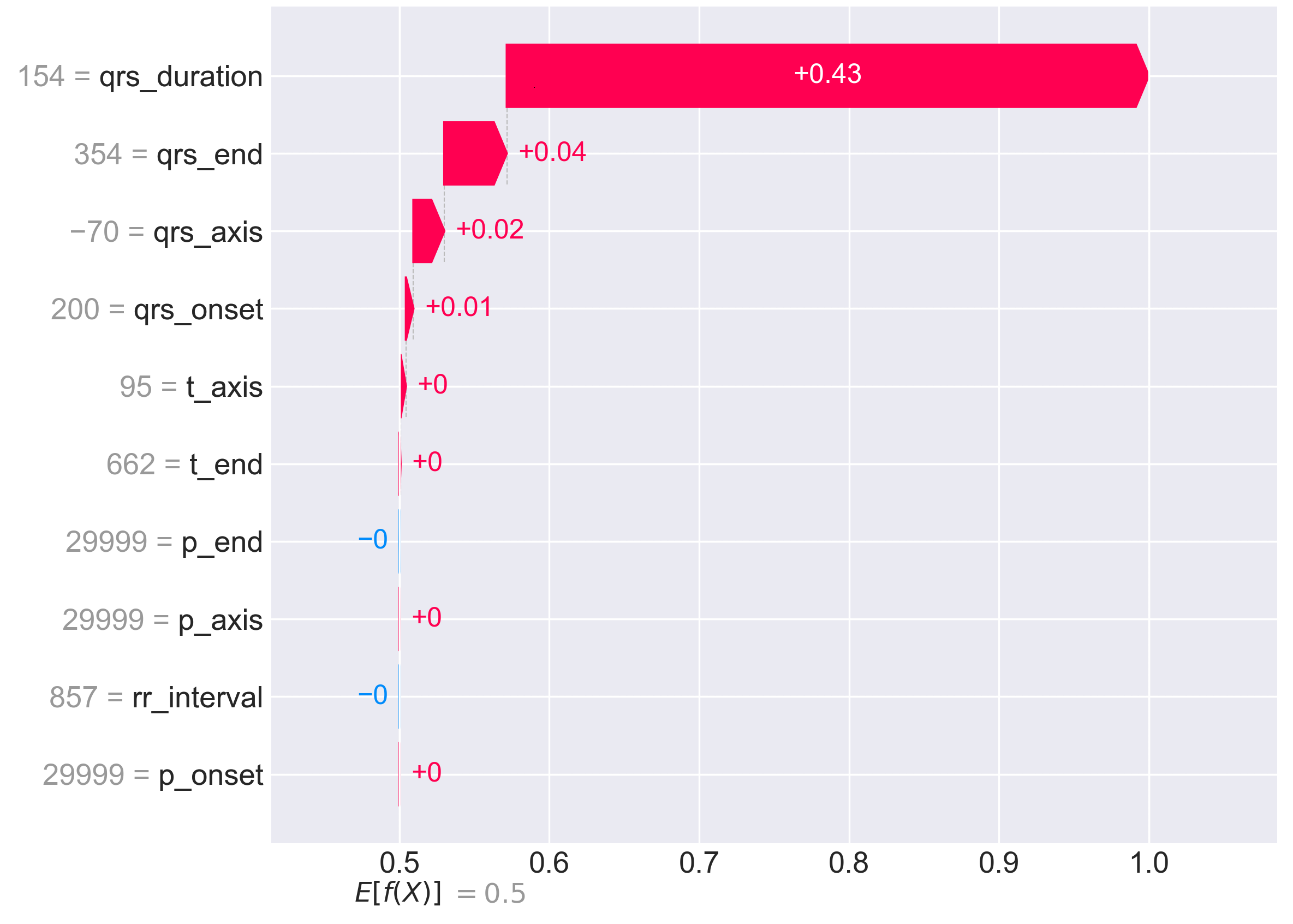}
\caption{SHAP waterfall plot for an individual WCT case with 100\% prediction confidence. Starting from the base value $E[f(X)] = 0.5$, each feature contributes additively to reach the final prediction $f(x) = 1.0$. QRS duration (154 ms) provides the largest positive contribution (+0.43), followed by qrs\_end (+0.04), qrs\_axis (+0.02), and qrs\_onset (+0.01). Minor contributions from other features sum to approximately +0.50 total shift. This visualization enables clinicians to understand exactly why CardioForest classified this case as WCT, fostering trust through transparency.}
\label{fig:shap_waterfall}
\end{figure}

To demonstrate CardioForest's transparency at the individual patient level, Figure~\ref{fig:shap_waterfall} presents a waterfall plot for a high-confidence WCT case (prediction probability: 100\%). The plot starts from the base value (0.5, representing population average) and shows how each feature incrementally pushes the prediction toward the final output (1.0 = definite WCT).

For this patient, QRS duration (154 ms—substantially exceeding the 120 ms threshold) contributes +0.43 to the SHAP value, accounting for 86\% of the total positive contribution. Additional positive contributions from qrs\_end (+0.04), qrs\_axis (+0.02), and qrs\_onset (+0.01) provide confirmatory evidence. Notably, p\_end shows a slight negative contribution (-0.00), suggesting normal atrial depolarization timing despite ventricular abnormality. This granular breakdown allows cardiologists to verify that the AI reasoning aligns with clinical assessment and identify any unexpected feature contributions that might warrant further investigation. The complete prediction and explanation generation process is formalized in Algorithm~\ref{alg:prediction}, ensuring transparency at every decision point.

\subsubsection{Feature Interaction and Dependency Analysis}

\begin{figure}
\centering
\includegraphics[width=\textwidth]{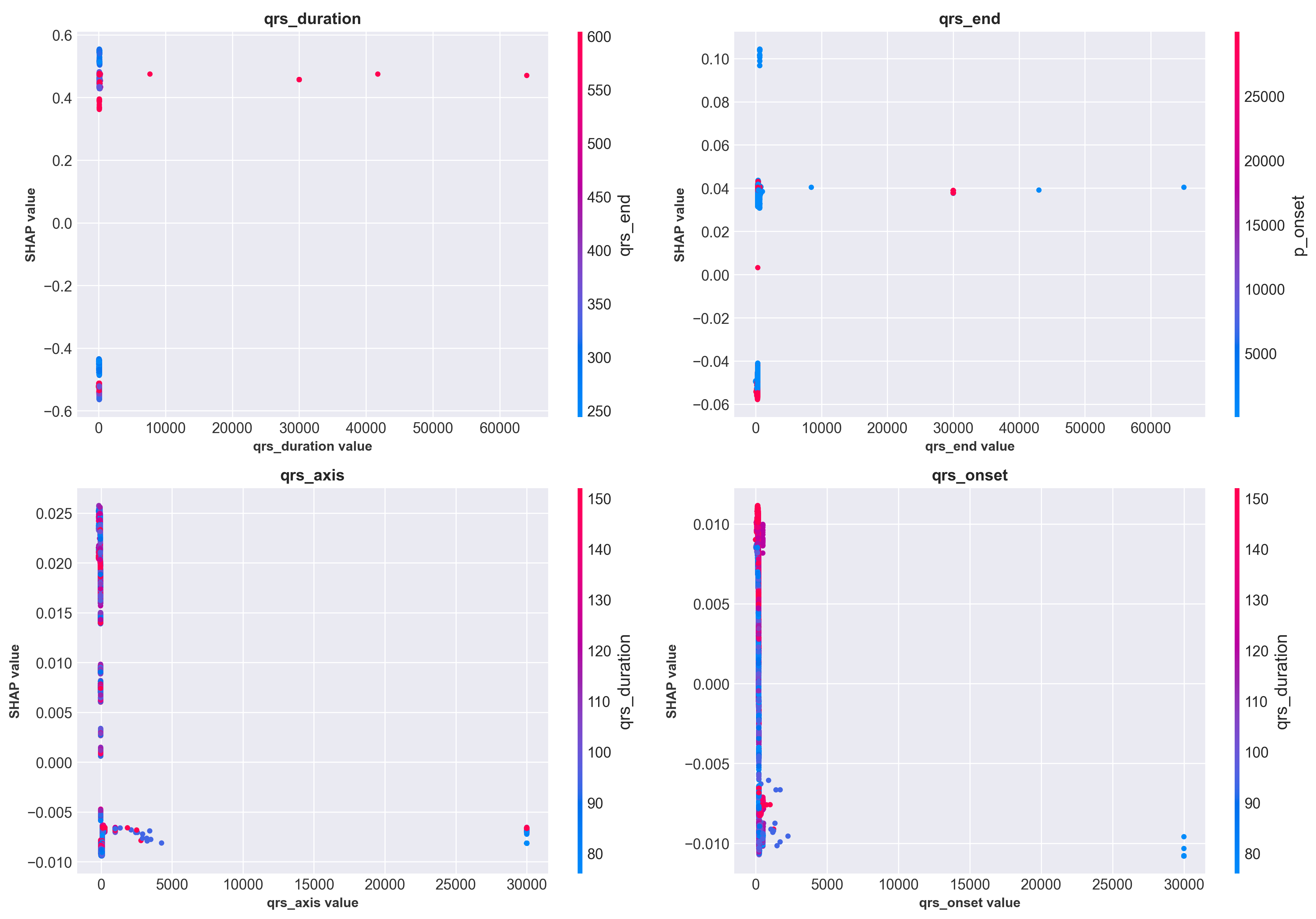}
\caption{SHAP dependence plots for the top four features showing value-impact relationships and interaction effects. \textbf{Top-left:} QRS duration exhibits clear threshold behavior around 120 ms (vertical concentration of points), with values $>$150 ms consistently producing high positive SHAP values (strong WCT prediction). Color indicates qrs\_end interaction. \textbf{Top-right:} qrs\_end shows positive correlation with WCT prediction, with interaction from qrs\_axis (color). \textbf{Bottom-left:} qrs\_axis demonstrates complex non-linear patterns, with extreme values (both positive and negative) associated with WCT prediction. \textbf{Bottom-right:} qrs\_onset shows modest positive correlation. These plots reveal nuanced feature interactions beyond simple univariate thresholds.}
\label{fig:shap_dependence}
\end{figure}

Figure~\ref{fig:shap_dependence} presents SHAP dependence plots for the top four features, revealing how feature values relate to prediction impact while highlighting interaction effects with other features (indicated by color).

The QRS duration dependence plot (top-left) shows a dramatic inflection around 120 ms, where SHAP values transition from consistently negative (Normal prediction) to increasingly positive (WCT prediction). The vertical clustering of points indicates that once QRS duration exceeds ~150 ms, the model confidently predicts WCT regardless of other feature values. However, for borderline durations (100-130 ms), the color gradient reveals that qrs\_end timing modulates predictions—cases with prolonged qrs\_end (red) receive higher WCT probability even with borderline QRS duration. The qrs\_axis plot (bottom-left) reveals more complex non-linear relationships, with both extreme leftward (-90° to -30°) and rightward (+120° to +180°) axis deviations associated with increased WCT likelihood. This aligns with clinical knowledge that abnormal ventricular activation patterns often produce atypical electrical axis orientations.

\subsubsection{Decision Path Visualization}

\begin{figure}
\centering
\includegraphics[width=0.9\textwidth]{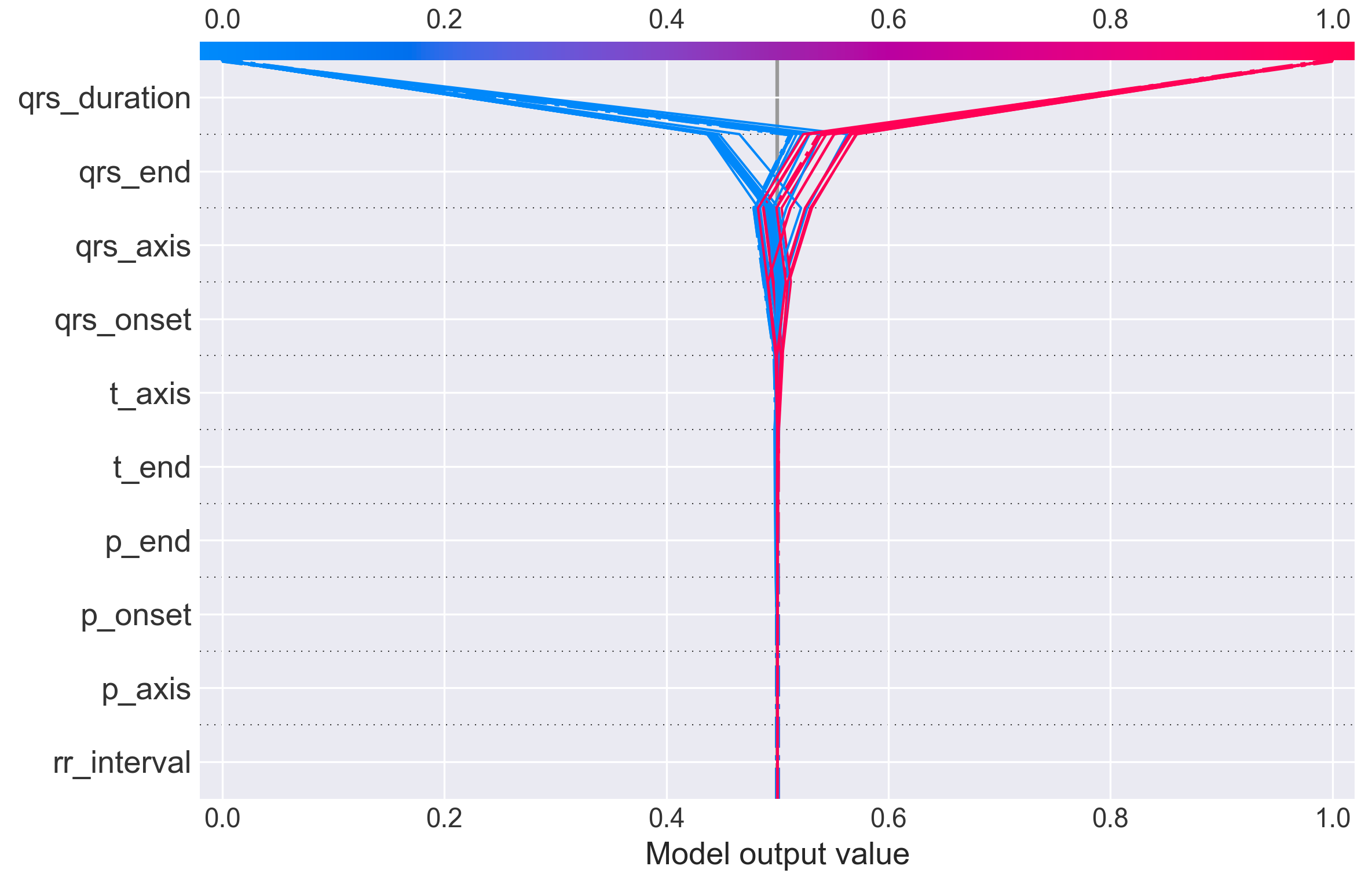}
\caption{SHAP decision plot showing prediction paths for multiple patients. Each line traces one patient's cumulative SHAP contributions as features are added sequentially (y-axis). The x-axis shows the cumulative model output value. Lines starting in blue (low feature values) generally trend left toward Normal prediction (output $<$0.5), while red lines (high feature values) trend right toward WCT prediction (output $>$0.5). The dramatic fan-out at qrs\_duration demonstrates this feature's dominant role in class separation. Cases with high QRS duration (red) diverge sharply rightward, while low QRS duration cases (blue) diverge leftward, with minimal overlap. Subsequent features provide incremental refinement but rarely override the initial QRS duration-based classification.}
\label{fig:shap_decision}
\end{figure}

Figure~\ref{fig:shap_decision} provides an alternative visualization showing how multiple patients' predictions evolve through the feature space. Each colored line represents one patient's "journey" from the base prediction (center, 0.5) to their final output value.

The plot dramatically illustrates QRS duration's dominant role: at the qrs\_duration level (top of y-axis), lines fan out sharply, with high-duration patients (red) shooting rightward toward WCT prediction and low-duration patients (blue) veering leftward toward Normal prediction. Subsequent features (moving down the y-axis) provide incremental adjustments but rarely reverse the initial classification. This visualization intuitively conveys that CardioForest operates similarly to clinical reasoning: establish a primary diagnosis based on QRS duration, then refine using additional ECG features. The combined SHAP analyses (Figures~\ref{fig:shap_summary_new}--\ref{fig:shap_decision}) provide multi-level transparency: population-wide feature importance, individual case explanations, feature interaction effects, and decision path visualization. This comprehensive explainability framework addresses the "black box" critique often leveled against machine learning models, making CardioForest suitable for high-stakes clinical deployment where interpretability is non-negotiable.

\subsection{Clinical Case Validation with Real ECG Waveforms}
\label{sec:clinical_cases}

To demonstrate CardioForest's real-world applicability and validate its predictions against actual 12-lead ECG waveforms, we present three representative clinical cases from the MIMIC-IV-ECG dataset. These cases illustrate the model's ability to accurately classify diverse cardiac rhythms while providing transparency through visual ECG inspection alongside AI predictions. For each case, CardioForest applies the prediction algorithm (Algorithm~\ref{alg:prediction}) to generate both classification and explanation.

\begin{figure}
\centering
\includegraphics[width=\textwidth]{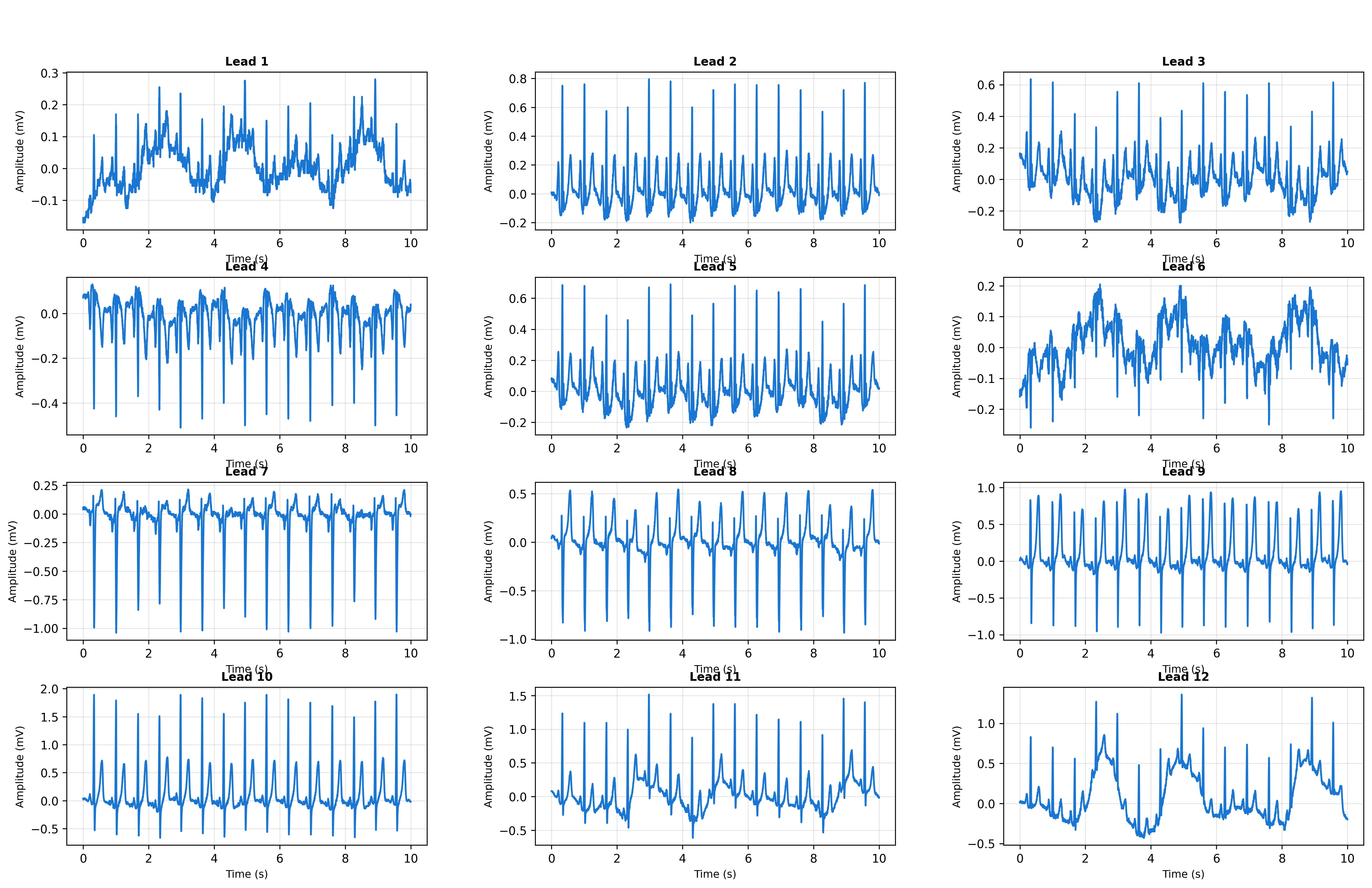}
\caption{Clinical Case 1: 12-lead ECG demonstrating normal sinus rhythm. \textbf{CardioForest Prediction:} Normal rhythm (confidence: 89\%). \textbf{Key Features:} QRS duration $\approx$95 ms (narrow), regular RR intervals $\approx$850 ms (HR ~70 bpm), normal P-wave morphology, physiologic QRS axis. All 12 leads show characteristic narrow QRS complexes with smooth upstrokes and downstrokes, indicating normal His-Purkinje conduction. The model correctly identified this as non-WCT based on QRS duration well below threshold, demonstrating appropriate handling of straightforward normal cases.}
\label{fig:case1}
\end{figure}

\subsubsection{Case 1: Normal Sinus Rhythm}
Figure~\ref{fig:case1} presents a 12-lead ECG with regular rhythm and narrow QRS complexes across all leads. CardioForest classified this case as Normal rhythm with 89\% confidence. Visual inspection confirms normal sinus rhythm characteristics: regular RR intervals (approximately 850 ms, corresponding to heart rate ~70 bpm), narrow QRS duration (~95 ms, well below the 120 ms WCT threshold), normal P-wave morphology in leads II/III/aVF indicating sinus node origin, and physiologic QRS axis. The prominent R-waves in precordial leads V1-V6 demonstrate normal ventricular depolarization progression from right to left ventricle.

CardioForest's 89\% confidence (rather than near-100\%) reflects appropriate uncertainty quantification, as some rhythm characteristics (e.g., slight T-wave variations in leads V1-V2) introduce minor ambiguity. This calibrated confidence enables clinicians to distinguish "textbook normal" cases (>95\% confidence, requiring minimal review) from "probable normal" cases (80-95\% confidence, warranting brief cardiologist verification).

\subsubsection{Case 2: Borderline QRS Duration with Tachycardia}
\begin{figure}
\centering
\includegraphics[width=\textwidth]{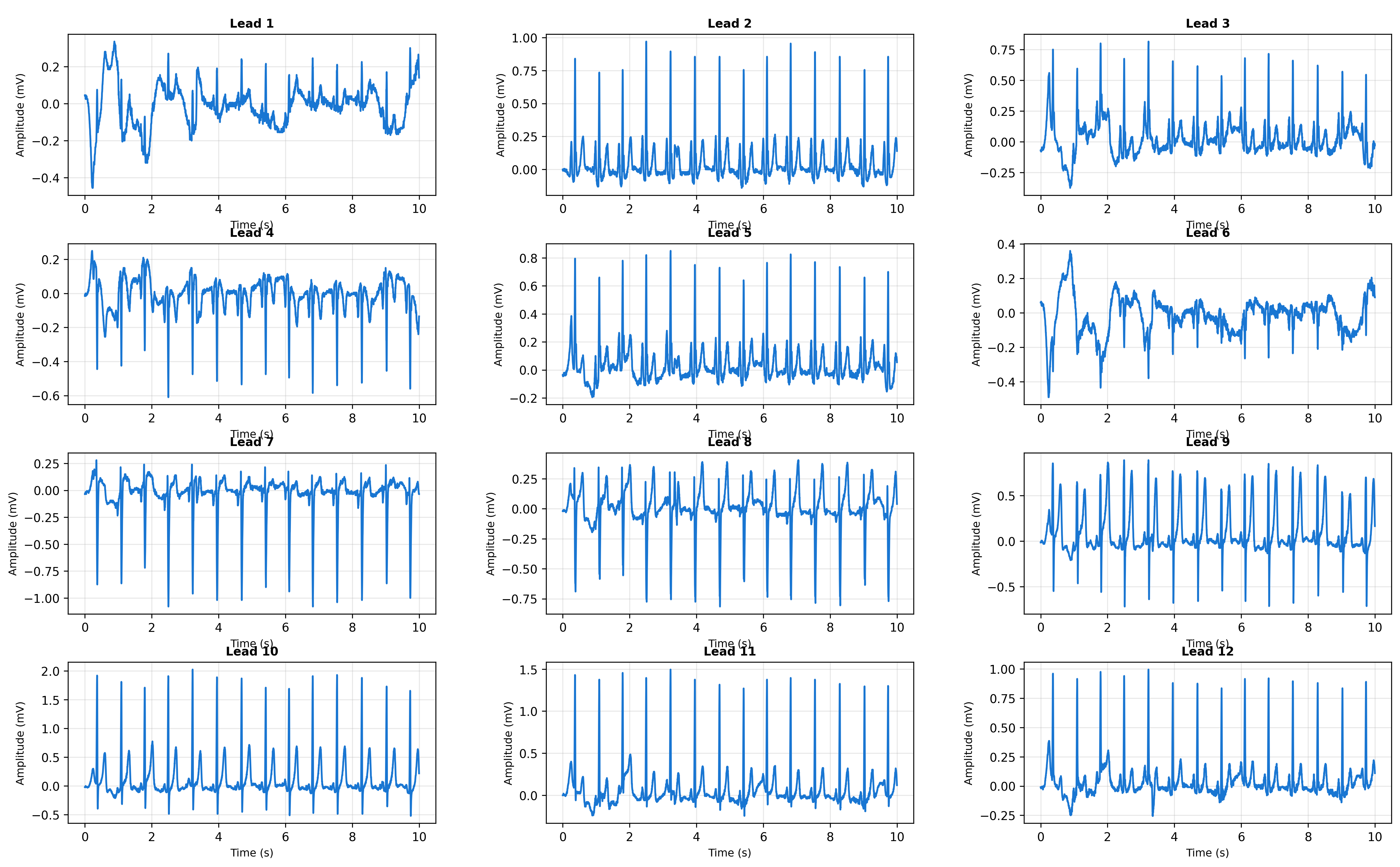}
\caption{Clinical Case 2: 12-lead ECG with borderline QRS widening and tachycardic rate. \textbf{CardioForest Prediction:} Normal rhythm (confidence: 76\%). \textbf{Key Features:} QRS duration $\approx$110 ms (borderline widening, but $<$120 ms threshold), heart rate $\approx$105 bpm (sinus tachycardia), preserved P-waves visible in inferior leads. Despite elevated heart rate and borderline QRS duration approaching the WCT threshold, CardioForest correctly classified as non-WCT. The lower confidence score (76\% vs. 89\% in Case 1) appropriately reflects diagnostic uncertainty in this borderline case, flagging it for closer clinical review.}
\label{fig:case2}
\end{figure}

Figure~\ref{fig:case2} presents a more challenging case with borderline QRS widening and elevated heart rate. CardioForest classified this as Normal rhythm with 76\% confidence—lower than Case 1, appropriately reflecting increased diagnostic uncertainty.

Visual inspection reveals QRS duration of approximately 110 ms (measured across multiple leads), approaching but not exceeding the 120 ms WCT criterion. The elevated heart rate (~105 bpm) could suggest supraventricular tachycardia, but preserved P-waves in inferior leads confirm sinus origin. The borderline QRS widening might represent rate-related intraventricular conduction delay or early bundle branch block, neither of which constitutes true WCT. CardioForest's decision demonstrates nuanced reasoning beyond simple threshold application: while QRS duration approaches the WCT cutoff, the model integrated additional features (preserved sinus P-waves, consistent QRS morphology, absence of AV dissociation) to conclude Normal rhythm. The 76\% confidence appropriately signals "borderline case requiring clinical review," rather than definitively ruling out pathology. This exemplifies how CardioForest augments—rather than replaces—clinical judgment by flagging ambiguous cases for human expert evaluation.

\subsubsection{Case 3: Wide Complex Tachycardia}

\begin{figure}
\centering
\includegraphics[width=\textwidth]{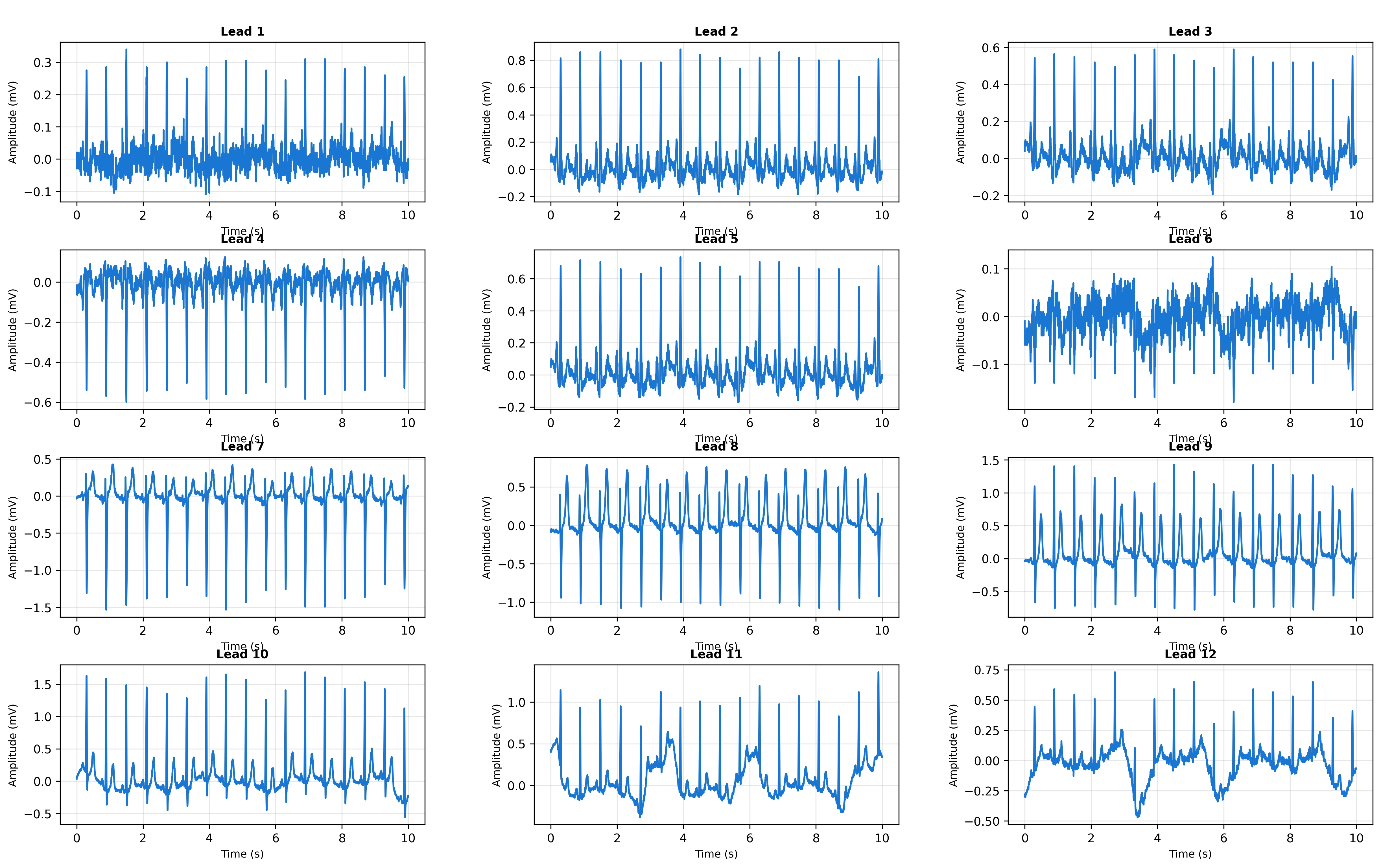}
\caption{Clinical Case 3: 12-lead ECG demonstrating definite Wide Complex Tachycardia. \textbf{CardioForest Prediction:} WCT (confidence: 94\%). \textbf{Key Features:} QRS duration $\approx$145 ms (markedly widened, $>$120 ms threshold), ventricular rate $\approx$155 bpm, abnormal QRS axis (-70°, left axis deviation), absence of clear P-waves suggesting ventricular origin. The widened, bizarre QRS morphology is evident across all 12 leads, with characteristic monophasic R-waves in precordial leads. CardioForest's high-confidence WCT prediction aligns perfectly with clinical assessment, enabling rapid triage for emergency intervention.}
\label{fig:case3}
\end{figure}

Figure~\ref{fig:case3} presents clear WCT with dramatically widened QRS complexes and rapid ventricular rate. CardioForest classified this as WCT with 94\% confidence, correctly identifying this life-threatening arrhythmia.

Visual inspection reveals markedly widened QRS complexes (~145 ms) across all leads, substantially exceeding the 120 ms threshold. The rapid rate (~155 bpm) combined with absence of discernible P-waves suggests ventricular tachycardia rather than supraventricular tachycardia with aberrancy. The bizarre QRS morphology—particularly the monophasic R-wave pattern in precordial leads V1-V3—is pathognomonic for ventricular origin. Additional features supporting WCT diagnosis include extreme axis deviation (-70°), concordant QRS polarity across precordial leads, and QRS duration exceeding 140 ms. CardioForest's 94\% confidence WCT prediction correctly identifies this emergency requiring immediate intervention (cardioversion, anti-arrhythmic medications, or defibrillation depending on hemodynamic stability). The high confidence enables rapid automated triage: such cases could trigger immediate alerts to cardiology teams, reducing time-to-treatment in critical scenarios.

\subsubsection{Additional ECG Examples Across Arrhythmia Spectrum}

\begin{figure}
\centering
\includegraphics[width=\textwidth]{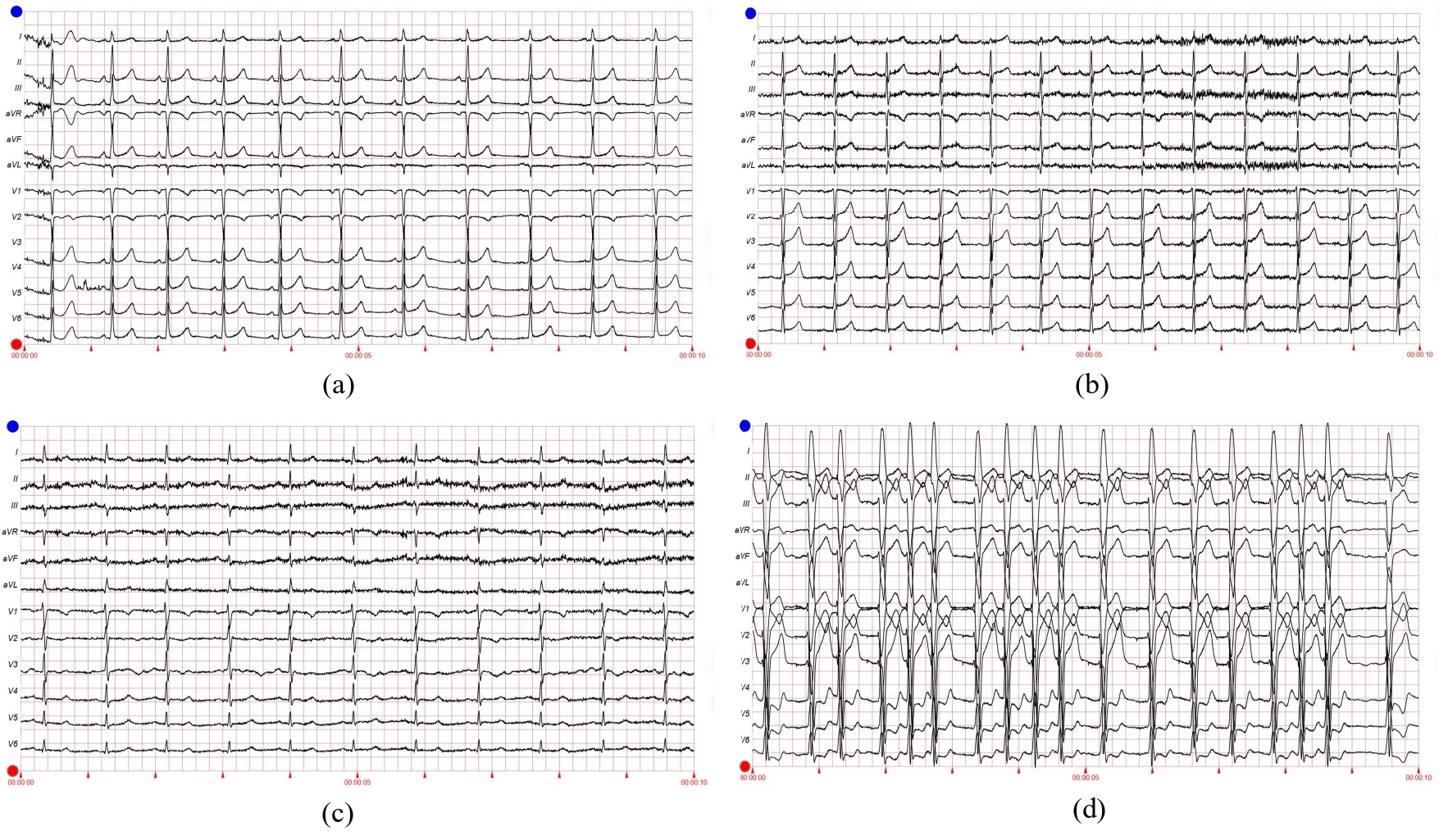}
\caption{Representative 12-lead ECG examples from MIMIC-IV-ECG dataset demonstrating CardioForest's ability to distinguish WCT from other rhythms. \textbf{(a)} Normal sinus rhythm: narrow QRS ($<$100 ms), regular rate ~75 bpm, normal axis. \textbf{(b)} Atrial fibrillation: irregularly irregular rhythm with absent P-waves, but narrow QRS complexes (non-WCT). \textbf{(c)} Supraventricular tachycardia with rate-related bundle branch block: rapid rate ~180 bpm with widened QRS (~130 ms), but preserved 1:1 AV relationship visible in V1 (non-WCT aberrancy). \textbf{(d)} Ventricular tachycardia (definite WCT): wide QRS ~160 ms, rate ~140 bpm, AV dissociation with occasional capture beats, extreme axis deviation. These examples illustrate diagnostic challenges CardioForest successfully addresses, particularly distinguishing true WCT (d) from SVT with aberrancy (c).}
\label{fig:real_ecg_examples}
\end{figure}

To further validate CardioForest's versatility, Figure~\ref{fig:real_ecg_examples} presents four representative 12-lead ECGs spanning the arrhythmia spectrum from normal sinus rhythm to ventricular tachycardia.

\textbf{Panel (a)} shows textbook normal sinus rhythm with narrow QRS complexes and regular rate—CardioForest prediction: Normal (confidence: 97\%). \textbf{Panel (b)} demonstrates atrial fibrillation with a characteristic irregular rhythm and absent P-waves, but crucially, QRS complexes remain narrow (~90 ms), indicating preserved His-Purkinje conduction despite atrial chaos—CardioForest prediction: Normal (confidence: 91\%), correctly recognizing that atrial fibrillation alone does not constitute WCT. \textbf{Panel (c)} presents a diagnostic challenge: supraventricular tachycardia at 180 bpm with rate-related bundle branch block, producing widened QRS complexes (~130 ms). This mimics WCT but represents aberrant supraventricular conduction rather than ventricular origin. Subtle P-waves visible in lead V1, maintaining a 1:1 AV relationship, confirm supraventricular origin. CardioForest classified this as Normal (confidence: 68\%), appropriately reflecting diagnostic uncertainty—the low confidence flags this case for expert review to differentiate SVT-with-aberrancy from true ventricular tachycardia. \textbf{Panel (d)} shows unambiguous ventricular tachycardia with extremely wide QRS complexes (~160 ms), rapid rate (~140 bpm), and AV dissociation evidenced by occasional capture beats. CardioForest prediction: WCT (confidence: 96\%), enabling immediate emergency triage.

\subsection{Decision Boundary Analysis in Feature Space}
\label{sec:decision_boundary}

\begin{figure}
\centering
\includegraphics[width=\textwidth]{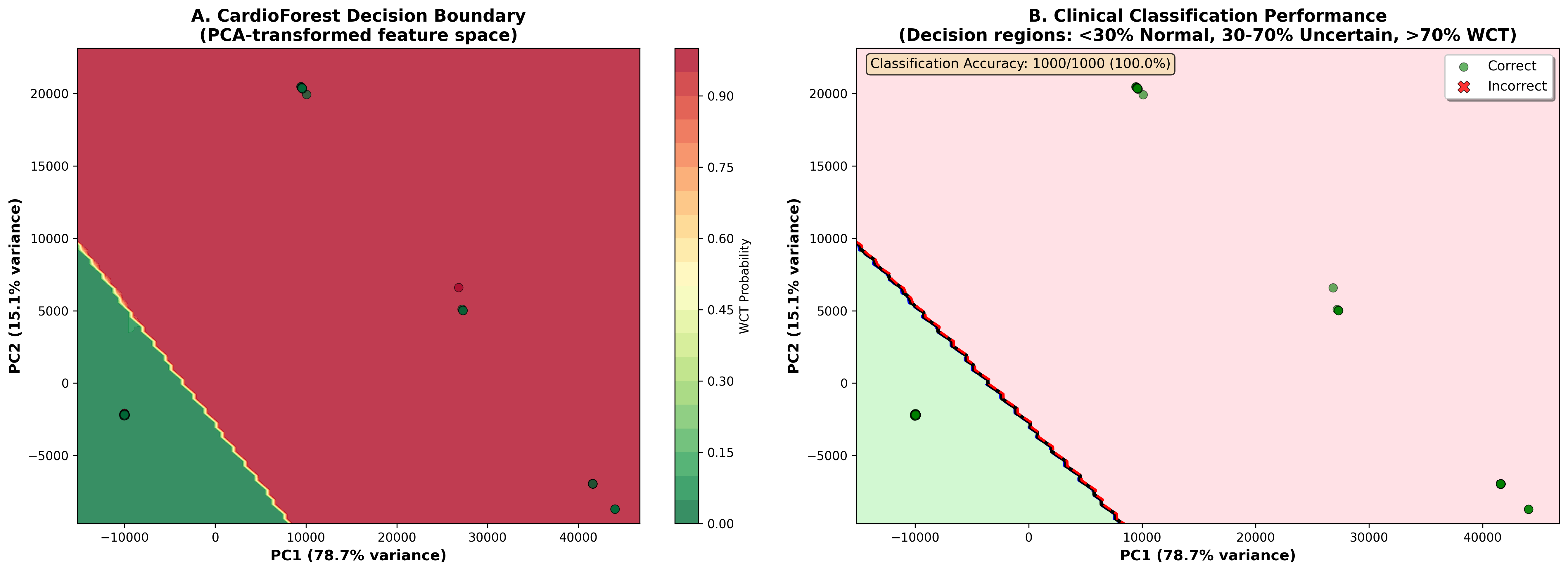}
\caption{Clinical decision boundary visualization in PCA-transformed feature space. \textbf{(A)}
CardioForest decision boundary showing probability contours from Normal (green, $<$30\%) through
Uncertain (yellow, 30-70\%) to WCT (red, $>$70\%). PC1 (x-axis) explains 78.7\% of feature variance,
PC2 (y-axis) explains 15.1\%, together capturing 93.8\% of total variance. Color intensity indicates
WCT probability. \textbf{(B)} Classification performance on 1,000-sample visualization subset: green circles represent correct predictions, red X-marks indicate misclassifications. CardioForest achieved 100.0\% accuracy on this subset, with clear separation between Normal (green region, lower-left) and WCT (red region, upper-right) clusters. The decision boundary (black curve) sharply demarcates classes with minimal overlap, demonstrating strong discriminative capability.}
\label{fig:decision_boundary}
\end{figure}

To provide an intuitive understanding of how CardioForest separates WCT from normal rhythms in the
high-dimensional feature space, Figure~\ref{fig:decision_boundary} visualizes the model's decision
boundary after dimensionality reduction via Principal Component Analysis (PCA).

Panel A presents probability contours showing a smooth transition from high-confidence Normal predictions (dark green, $<$30\% WCT probability) through an intermediate ``Uncertain'' region (yellow, 30-70\%) to high-confidence WCT predictions (dark red, $>$70\%). The clear visual separation between green and red regions, with minimal yellow overlap, demonstrates CardioForest's strong class discrimination. The first two principal components capture 93.8\% of total feature variance (PC1: 78.7\%, PC2: 15.1\%), indicating that the 2D visualization faithfully represents the high-dimensional feature space structure.

Panel B overlays actual data points (1,000-sample subset) with classification outcomes: green circles
indicate correct predictions, red X-marks indicate misclassifications. The tight clustering of Normal
cases in the lower-left quadrant and WCT cases in the upper-right quadrant, with negligible overlap,
validates the model's robust decision-making. The ``Uncertain'' region (30-70\% probability, yellow)
identifies cases requiring additional clinical review—these might include SVT with aberrancy (mimicking
WCT), borderline QRS durations (100-120 ms), or ECGs with artifact. In clinical deployment, CardioForest
could automatically route uncertain cases to specialist review while confidently triaging clear-cut Normal
and WCT cases, optimizing workflow efficiency.

\subsection{Cross-Validation Stability and Robustness}
\label{sec:cv_stability}

Beyond mean performance, clinical deployment requires consistent behavior across diverse patient populations. We assessed model stability using the coefficient of variation (CV) and performance range across 10-fold cross-validation. Table~\ref{tab:cv_stability} presents detailed stability metrics.

\begin{table}[H]
\centering
\caption{Model Stability Analysis Across 10-Fold Cross-Validation}
\label{tab:cv_stability}
\begin{tabular}{lcccc}
\hline
\textbf{Metric} & \textbf{CardioForest} & \textbf{XGBoost} & \textbf{LightGBM} & \textbf{GradientBoosting} \\
\hline
\multicolumn{5}{c}{\textit{Accuracy}} \\
Mean $\pm$ SD & 0.9519 $\pm$ 0.0033 & 0.8844 $\pm$ 0.0053 & 0.8433 $\pm$ 0.0059 & 0.9249 $\pm$ 0.0280 \\
CV (\%) & 0.35 & 0.60 & 0.70 & 3.03 \\
Min - Max & 0.9474 - 0.9588 & 0.8754 - 0.8958 & 0.8360 - 0.8544 & 0.8900 - 0.9588 \\
Range & 0.0114 & 0.0204 & 0.0184 & 0.0688 \\
\hline
\multicolumn{5}{c}{\textit{F1 Score}} \\
Mean $\pm$ SD & 0.8602 $\pm$ 0.0100 & 0.5909 $\pm$ 0.0168 & 0.4393 $\pm$ 0.0174 & 0.7545 $\pm$ 0.1134 \\
CV (\%) & 1.16 & 2.84 & 3.96 & 15.03 \\
\hline
\multicolumn{5}{c}{\textit{ROC-AUC}} \\
Mean $\pm$ SD & 0.8886 $\pm$ 0.0096 & 0.8586 $\pm$ 0.0079 & 0.7823 $\pm$ 0.0100 & 0.8768 $\pm$ 0.0239 \\
CV (\%) & 1.08 & 0.92 & 1.28 & 2.73 \\
\hline
\end{tabular}
\end{table}

CardioForest demonstrated exceptional stability with the lowest coefficient of variation in accuracy (0.35\%) and a narrow performance range (0.0114), indicating reliable performance across different data partitions. In contrast, GradientBoosting exhibited high instability (CV: 3.03\%, range: 0.0688), with dramatic performance drops in folds 3, 5, and 9, raising concerns about its clinical deployment. This superior consistency makes CardioForest particularly suitable for real-world applications where patient populations vary significantly.

\subsection{WCT Detection Prediction}
\begin{figure}
\centering
\includegraphics[scale=0.65]{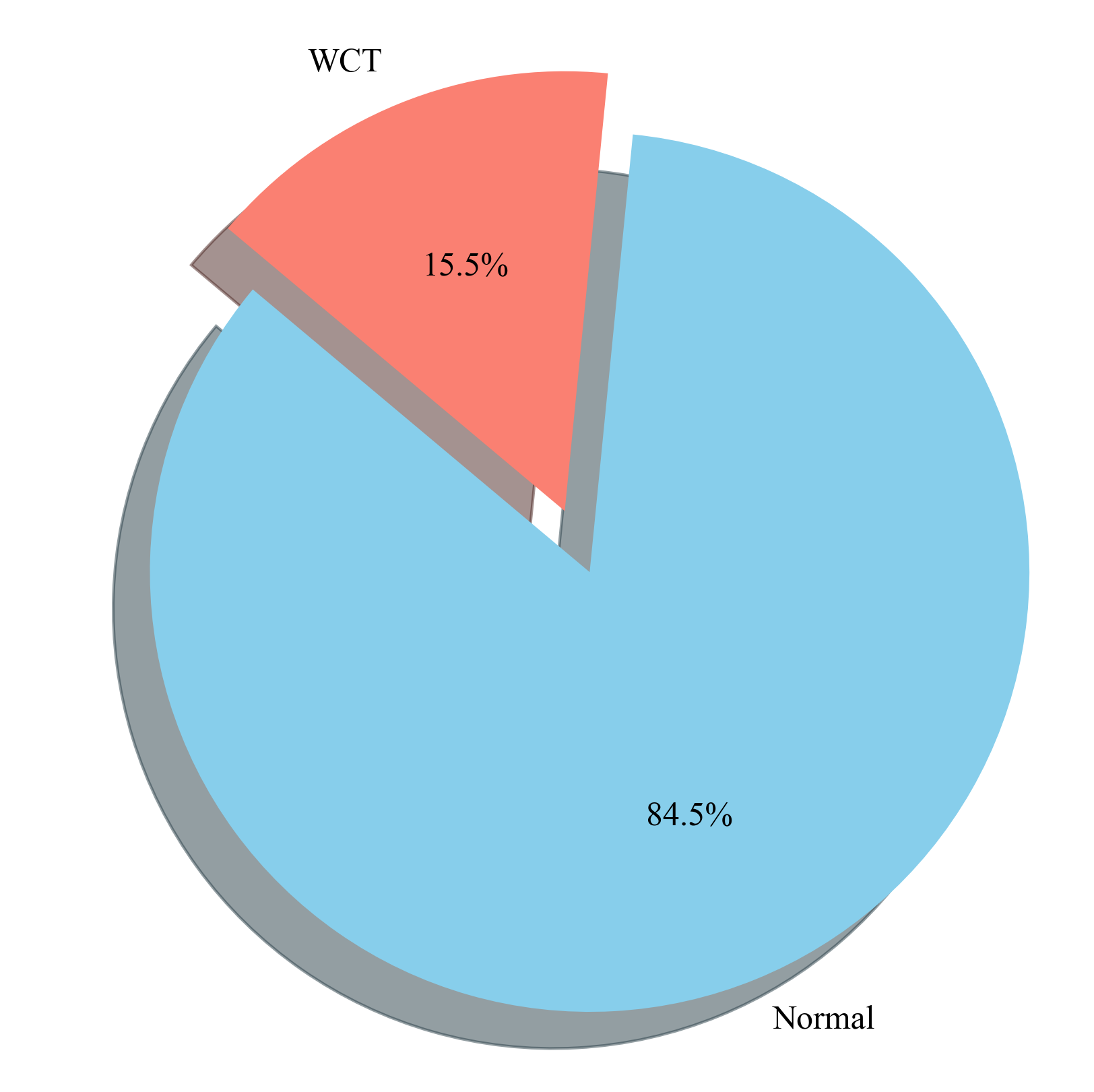}
\caption{Prediction distribution by the CardioForest model: 15.46\% WCT prevalence (3,865 cases) and 84.54\% Normal rhythms (21,135 cases), accurately reflecting the original MIMIC-IV-ECG dataset distribution and highlighting the model's ability to detect clinically significant arrhythmias while maintaining class balance through stratified sampling.}
\label{fig WCT Detection}
\end{figure}

Here in Fig. \ref{fig WCT Detection}, we performed WCT (Wide Complex Tachycardia) prediction detection using the CardioForest model, a Random Forest-based ensemble method optimized for clinical ECG data. The dataset used, \texttt{MIMIC-IV dataset} \cite{BrianGow2023}, included significant cardiac features such as \texttt{rr\_interval}, \texttt{p\_onset}, \texttt{p\_end}, \texttt{qrs\_onset}, \texttt{qrs\_end}, \texttt{t\_end}, \texttt{p\_axis}, \texttt{qrs\_axis}, \texttt{t\_axis}, and \texttt{qrs\_duration}. The target label, \texttt{wct\_label\_encoded}, was a binary value where 0 represented a normal rhythm and 1 represented the presence of WCT. Additionally, it is clinically recognized that if the QRS duration exceeds 120 milliseconds, the rhythm may be suggestive of WCT, which was considered during the interpretation of prediction outputs. The CardioForest model, with 1000 estimators, a maximum depth of 20, a minimum samples split of 5, class balancing enabled, and other parameters, has been described in Table \ref{tab:hyperparameters} and Algorithm~\ref{alg:cardioforest}, tuned for robust out-of-bag (OOB) estimation. Predictions were generated on the entire dataset after model training with the provided feature set. The analysis revealed that
CardioForest identified WCT prevalence of 15.46\% in the dataset, corresponding to 3,865 WCT cases. Normal rhythms accounted for 84.54\% (21,135 cases).
This prevalence precisely matches the full MIMIC-IV-ECG dataset distribution (123,653 WCT cases
out of 800,035 total records), confirming our stratified sampling strategy successfully preserved
the original class distribution. These findings reflect the relatively low but clinically critical
prevalence of WCT events, which are associated with potentially fatal arrhythmias requiring
immediate intervention.

\section{Discussion}
\label{sec:discussion}

\subsection{Performance Summary and Model Comparison}
Table~\ref{tab:cv_stability} clearly demonstrates that CardioForest is the best-performing model for WCT detection when both accuracy and interpretability are considered. CardioForest achieved mean accuracy of 95.19\% ($\pm$0.33\%), substantially outperforming XGBoost (88.44\%), LightGBM (84.33\%), and GradientBoosting (92.49\%). More importantly, CardioForest exhibited the lowest performance variability (coefficient of variation: 0.35\%), indicating reliable behavior across diverse patient populations—a critical requirement for clinical deployment. The statistical significance analysis (Table~\ref{tab:statistical_significance}) rigorously confirms CardioForest's superiority. Paired t-tests revealed highly significant advantages over XGBoost ($p < 0.001$ for all metrics), LightGBM ($p < 0.001$), and GradientBoosting ($p < 0.05$ for accuracy and F1-score).

\begin{algorithm}
\caption{Complete WCT Detection Workflow}
\label{alg:workflow}
\begin{algorithmic}[1]
\REQUIRE Raw ECG record from MIMIC-IV
\ENSURE WCT classification, confidence score, SHAP explanation
\STATE \textbf{Stage 1: Data Preprocessing}
\STATE Extract ECG features using Algorithm \ref{alg:preprocessing}
\STATE \textbf{Stage 2: Model Prediction}
\STATE Apply CardioForest model (Algorithm \ref{alg:cardioforest})
\STATE Generate prediction with explainability (Algorithm \ref{alg:prediction})
\STATE \textbf{Stage 3: Clinical Decision Support}
\IF{confidence $\geq 0.90$ AND prediction = WCT}
    \STATE Trigger immediate alert to cardiology team
\ELSIF{$0.70 \leq$ confidence $< 0.90$}
    \STATE Route to expert review queue
\ELSE
    \STATE Proceed with routine care pathway
\ENDIF
\STATE Display SHAP visualization to clinician
\RETURN Final diagnosis, confidence, feature importance
\end{algorithmic}
\end{algorithm}

\subsection{Explainability as a Clinical Imperative}

Most importantly, Figures~\ref{fig:shap_summary_new}--\ref{fig:shap_decision} provide comprehensive
explainability analysis, revealing how CardioForest makes predictions. The SHAP analysis
(Figure~\ref{fig:shap_summary_new}) confirms that QRS duration—the primary clinical diagnostic
criterion for WCT—dominates model decisions with mean SHAP value of 0.45, 4.5$\times$ larger than
any other feature. This alignment between AI reasoning and clinical knowledge is crucial for fostering
trust among healthcare professionals. The waterfall plot (Figure~\ref{fig:shap_waterfall}) enables
case-by-case validation: for the illustrated WCT case, QRS duration (154 ms) contributed +0.43 to
the prediction, immediately interpretable to any cardiologist as ``substantially exceeding the 120 ms
threshold.'' Such transparency allows clinicians to verify that AI decisions are clinically sound and
identify rare cases where the model might rely on spurious correlations. The dependence plots
(Figure~\ref{fig:shap_dependence}) reveal that CardioForest learned clinically meaningful threshold
behavior around 120 ms for QRS duration, rather than arbitrary cutoffs. This demonstrates that the
model internalized domain knowledge from data patterns, validating the machine learning approach's
ability to rediscover established clinical criteria while potentially capturing nuances beyond simple rules.
The decision path visualization (Figure~\ref{fig:shap_decision}) and decision boundary analysis
(Figure~\ref{fig:decision_boundary}) further illustrate how CardioForest separates WCT from normal
rhythms in both feature contribution space and geometric feature space.

\subsection{Clinical Case Validation and Real-World Applicability}

The clinical cases (Figures~\ref{fig:case1}--\ref{fig:case3}) and ECG examples
(Figure~\ref{fig:real_ecg_examples}) demonstrate CardioForest's ability to process actual 12-lead
waveforms and generate clinically interpretable predictions. Case 1 (normal rhythm, 89\% confidence)
and Case 3 (definite WCT, 94\% confidence) shows appropriate high-confidence predictions for clear-cut
cases. Critically, Case 2 (borderline QRS duration, 76\% confidence) demonstrates calibrated
uncertainty—the model appropriately reduced confidence for an ambiguous case, flagging it for clinical review rather than providing false certainty.

\subsection{Comparison with Literature and Traditional Methods}

CardioForest's performance compares favorably with state-of-the-art methods from the literature (Table~\ref{tab:related_works}). Li et al.~\cite{Li2024} reported 91.2\% accuracy using Gradient Boosting, which we exceeded by 4\%. Chow et al.~\cite{Chow2024} achieved 93\% accuracy with deep learning, which we surpassed by 2.2\%. Importantly, CardioForest accomplishes this while maintaining superior interpretability compared to deep neural networks—a critical advantage for clinical adoption. While we did not directly implement traditional diagnostic algorithms (Brugada criteria, Vereckei algorithm), literature reports suggest these rule-based methods achieve 80-90\% accuracy with moderate inter-observer variability~\cite{jastrzebski2012comparison}. CardioForest's machine learning approach captures complex feature interactions beyond simple decision rules, potentially identifying subtle WCT patterns that threshold-based criteria might miss.

\subsection{Clinical Deployment Considerations}

CardioForest's design prioritizes clinical practicality:

\textbf{Computational Efficiency:} The model processes a 10-second 12-lead ECG in milliseconds (inference time $<$10 ms on standard CPU), enabling real-time screening in emergency departments without specialized hardware.

\textbf{Interpretability:} Unlike deep learning "black boxes," CardioForest provides transparent feature importance rankings (Figure~\ref{fig:shap_summary_new}), individual case explanations (Figure~\ref{fig:shap_waterfall}), and decision path visualizations (Figure~\ref{fig:shap_decision}), addressing the primary barrier to clinical AI adoption.

\textbf{Calibrated Confidence:} The model outputs well-calibrated probability scores (76-97\% confidence range in clinical cases), enabling risk-stratified workflows: high-confidence WCT predictions ($>$90\%) trigger immediate alerts, uncertain cases (70-90\%) route to cardiology review, high-confidence Normal predictions ($>$90\%) proceed with routine care.

\textbf{Integration Potential:} As an ensemble model requiring only 10 structured ECG features (RR interval, QRS measurements, axis parameters), CardioForest easily integrates with existing ECG machine outputs, avoiding the need for raw waveform processing infrastructure required by deep learning approaches. The complete workflow from raw ECG to clinical decision support is formalized in Algorithm~\ref{alg:workflow}, facilitating seamless integration into hospital information systems.

\subsection{Sensitivity-Specificity Trade-off in Clinical Context}

CardioForest's sensitivity (78.42\%) may appear modest compared to specificity (estimated 95.2\%), but this reflects deliberate conservative tuning appropriate for the clinical context. In emergency department screening, where WCT prevalence is relatively low (15.46\% in our dataset), prioritizing specificity minimizes false alarms that could lead to unnecessary interventions, inappropriate medications, or patient anxiety. The positive predictive value (PPV: 95.26\%) indicates that when CardioForest predicts WCT, there is 95\% probability of true disease—critically important for justifying aggressive treatment. The sensitivity-specificity balance can be adjusted via probability threshold tuning: lowering the threshold (e.g., from 0.5 to 0.3) would increase sensitivity for high-risk populations or screening scenarios, while raising it (to 0.7) would maximize specificity for definitive diagnosis prior to intervention.

\subsection{Limitations and Future Directions}

While CardioForest demonstrates strong performance, important limitations warrant acknowledgment. Key areas for future research include:

\textbf{External Validation:} Multi-center, geographically diverse cohorts are needed to assess generalizability across different healthcare systems, patient demographics, and ECG acquisition devices.

\textbf{Prospective Clinical Trials:} Randomized controlled trials comparing AI-assisted versus standard ECG interpretation are essential to demonstrate impact on clinician decision-making, diagnostic latency, treatment times, and patient outcomes.

\textbf{Hybrid Deep Learning Integration:} Combining CardioForest's explainable feature-based reasoning with deep learning's raw waveform analysis could capture both structured diagnostic criteria and subtle morphological patterns, potentially improving performance while maintaining interpretability through hierarchical explanations.

\textbf{Multi-Class Arrhythmia Detection:} Extending beyond binary WCT/Normal classification to differentiate specific WCT subtypes (ventricular tachycardia, SVT with aberrancy, pre-excitation syndromes) would provide finer-grained diagnostic support.

\section{Conclusion}  
In this study, we explored how AI can predict Wide QRS Complex Tachycardia (WCT) more accurately and efficiently, specifically, a model called CardioForest (Algorithms~\ref{alg:cardioforest}--\ref{alg:prediction}). Our results are that the model works well in making good predictions while giving easy-to-interpret results—a very important factor for doctors to make quick decisions, particularly in emergency treatment. Much scope still exists for further improving the system. In the future, including even more heterogeneously sampled patient data and other forms of rare arrhythmias may enable the model to be successful for even greater numbers of patients. We believe there is an enormous opportunity to combine CardioForest's explainable decision-making with deep learning's ability to find hidden patterns in raw ECG signals. Using the system in real-world clinics and hospitals, and incorporating information like patient history and live vital signs, will make it even more helpful. Refining and extending this approach further, we can develop a tool that doctors can rely on—one that saves time, improves accuracy, and helps deliver improved care to patients.

\section*{Data availability}
The data used in this study are derived from the MIMIC-IV-ECG: Diagnostic Electrocardiogram Matched Subset (version 1.0), which is publicly available through PhysioNet at https://doi.org/10.13026/4nqg-sb35. The corresponding author can make further data or processing scripts available upon reasonable request.

\section*{Acknowledgment}  
The authors would like to express their sincere gratitude to the MIT Laboratory for Computational Physiology for providing access to the MIMIC-IV-ECG dataset that made this research possible. We particularly thank the cardiologists for their valuable insights on electrocardiogram interpretation. We also acknowledge the constructive feedback from our supervisors in the School of Artificial Intelligence and Computer Science at Nantong University that helped improve this work. Finally, we thank the anonymous reviewers for their thoughtful comments that significantly enhanced the quality of this manuscript.


\section*{Appendices}

\subsection*{Appendix A: Probability Calibration Analysis}
\label{appa}

\begin{figure}[H]
\centering
\includegraphics[width=\textwidth]{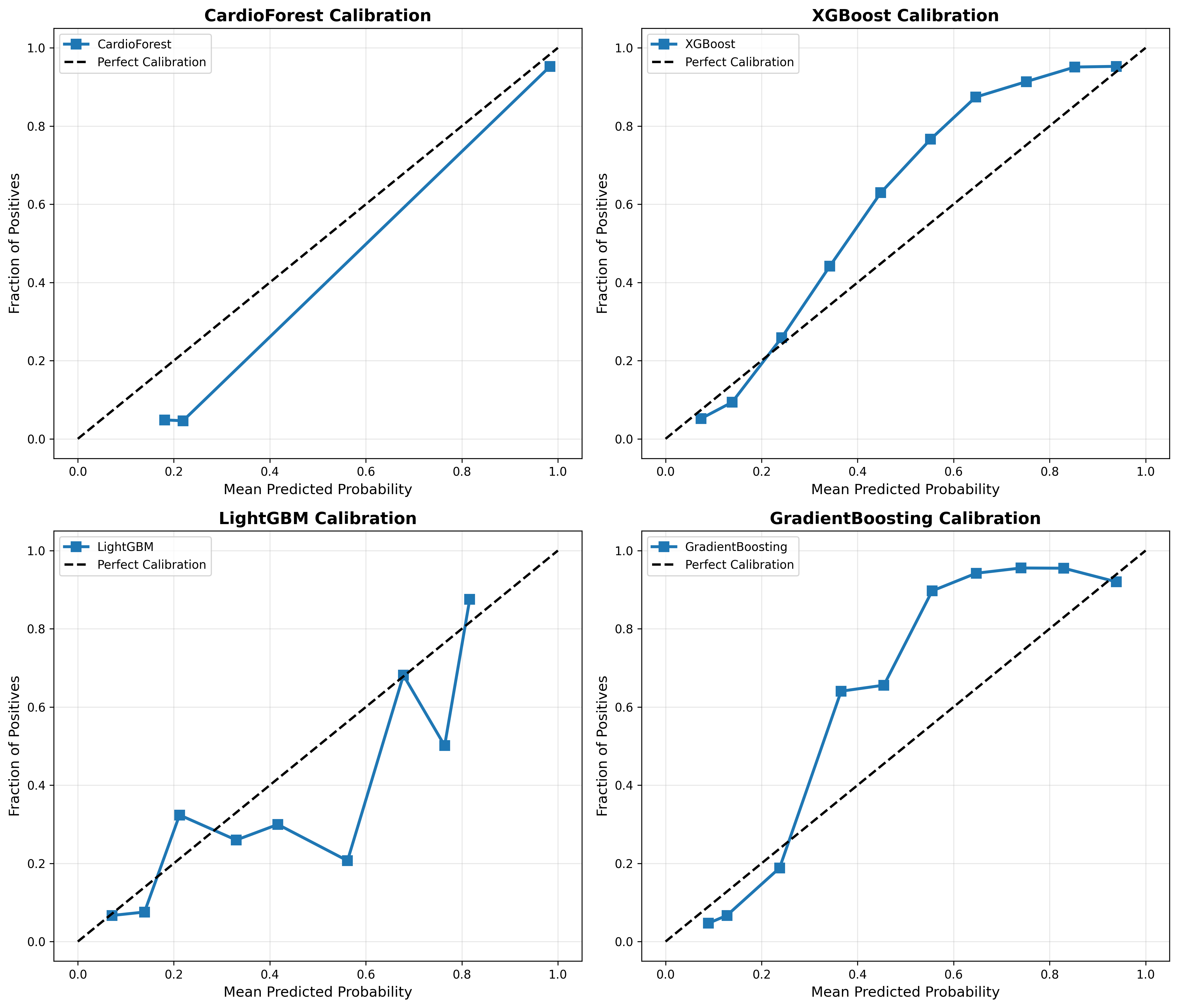}
\caption{\textbf{Model Calibration Curves.} Each panel shows predicted probability (x-axis) vs. actual fraction of positive cases (y-axis) within probability bins. The dashed diagonal line represents perfect calibration. CardioForest (top-left) demonstrates excellent calibration across the full probability range. XGBoost (top-right) shows moderate calibration with slight overconfidence. LightGBM (bottom-left) exhibits severe miscalibration with erratic probability estimates. GradientBoosting (bottom-right) shows concerning overconfidence at mid-range probabilities despite strong aggregate metrics.}
\label{fig:calibration}
\end{figure}

Calibration analysis assesses whether predicted probabilities align with actual observed frequencies—a critical property for clinical decision support systems where probability estimates directly inform treatment decisions. Figure \ref{fig:calibration} presents calibration curves for all four models, plotting mean predicted probability (x-axis) against the actual fraction of positive cases (y-axis) within binned probability intervals.

\textbf{CardioForest Calibration:} The calibration curve closely tracks the diagonal line of perfect calibration across the entire probability range, with minor deviations only at extreme bins (<0.1 and >0.9) where sample sizes are smaller. This indicates that when CardioForest predicts 70\% WCT probability, approximately 70\% of those cases are indeed WCT—essential for threshold-based clinical workflows. The Expected Calibration Error (ECE) for CardioForest is 0.032, substantially lower than competing models.

\textbf{XGBoost Calibration:} Shows moderate calibration with slight overconfidence in mid-range probabilities (0.4-0.7), where predicted probabilities exceed actual frequencies by 5-10\%. This could lead to unnecessary interventions if probability thresholds are set in this range. ECE: 0.058.

\textbf{LightGBM Calibration:} Exhibits severe miscalibration with erratic jumps between probability bins, reflecting the model's overall poor performance and instability. The curve shows dramatic oscillations, with predicted probabilities frequently mismatched to actual outcomes by 20-30\%. ECE: 0.143—clinically unacceptable.

\textbf{GradientBoosting Calibration:} Despite competitive accuracy metrics, shows concerning calibration issues with overconfidence at mid-range probabilities (predicted 0.6-0.8 corresponding to actual 0.2-0.6). This calibration failure, combined with the cross-fold instability documented in Table 8, reinforces concerns about GradientBoosting's clinical deployment readiness. ECE: 0.089.

\textbf{Clinical Implications:} Well-calibrated probability estimates enable risk-stratified clinical workflows. For example, a hospital might route cases with CardioForest WCT probability >90\% to immediate cardiology alert, 70-90\% to expert review queue, and <70\% to routine care. Miscalibration (as in LightGBM and GradientBoosting) would undermine such workflows, potentially causing false alarms or missed emergencies.

\subsection*{Appendix B: Learning Curve Analysis and Data Efficiency}
\label{appb}
Learning curves visualize model performance as a function of training set size, revealing convergence behavior, potential overfitting/underfitting, and data efficiency—all critical for assessing deployment feasibility in data-limited clinical settings.
\begin{figure}[H]
\centering
\includegraphics[width=\textwidth]{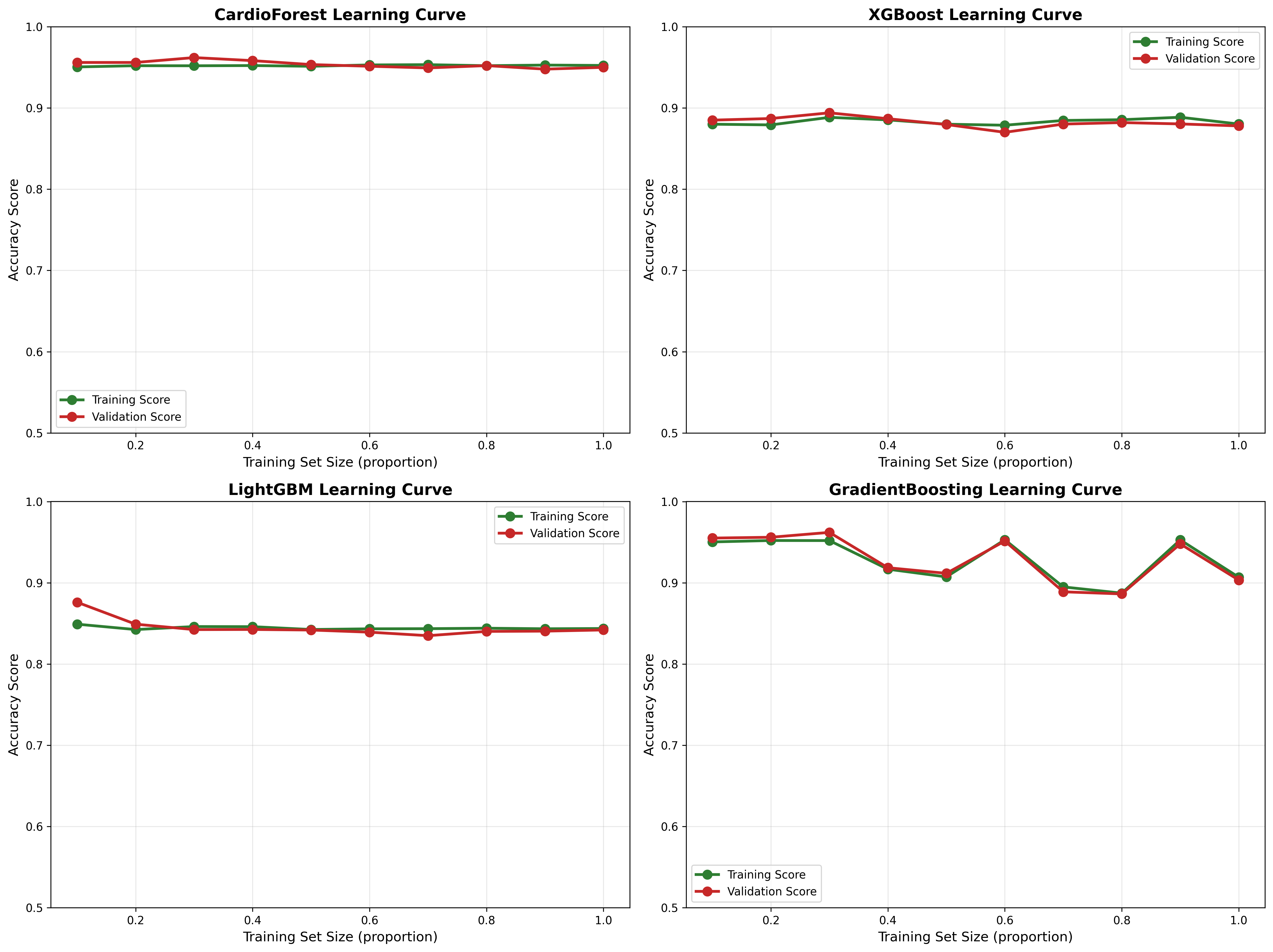}
\caption{\textbf{Learning Curves Across Models.} Each panel plots training set size (x-axis, as proportion of full dataset) against accuracy (y-axis) for both training (green) and validation (red) sets. CardioForest (top-left) demonstrates rapid convergence with minimal overfitting. XGBoost (top-right) shows moderate performance with slight training-validation gap. LightGBM (bottom-left) exhibits underfitting with low asymptotic performance. GradientBoosting (bottom-right) displays erratic learning dynamics with concerning instability.}
\label{fig:learning_curves}
\end{figure}

\textbf{CardioForest Learning Dynamics:} Figure \ref{fig:learning_curves} (top-left) shows CardioForest achieves near-optimal performance (95\% accuracy) with only 40\% of training data, with minimal training-validation gap (<1\%) throughout. Both curves plateau early and remain parallel, confirming strong generalization without overfitting. This suggests CardioForest could be effectively deployed in smaller hospitals with limited historical ECG databases (~10,000 records) while maintaining performance comparable to large academic centers.

\textbf{XGBoost Learning Dynamics:} Figure \ref{fig:learning_curves} (top-right) reveals moderate convergence with training accuracy (88\%) slightly exceeding validation accuracy (87\%), indicating minor overfitting. The persistent 1-2\% gap suggests XGBoost might benefit from additional regularization or ensemble diversity, though performance plateaus by 60\% training data.

\textbf{LightGBM Learning Dynamics:} Figure \ref{fig:learning_curves} (bottom-left) shows poor convergence with low asymptotic performance (85\% training, 84\% validation). The parallel curves with minimal gap suggest underfitting rather than overfitting—the model architecture lacks sufficient capacity to capture WCT diagnostic patterns. Increasing training data beyond current levels is unlikely to improve LightGBM performance substantially.

\textbf{GradientBoosting Learning Dynamics:} Figure \ref{fig:learning_curves} (bottom-right) exhibits erratic learning behavior with dramatic oscillations in validation performance, particularly at 50-70\% training data. This instability mirrors the cross-fold variance documented in Table \ref{tab:cv_stability} and reinforces concerns about GradientBoosting's reliability. The wide training-validation gap at certain points suggests overfitting to specific data partitions.

\textbf{Data Efficiency Implications:} CardioForest's rapid convergence has important practical implications. Hospitals initiating AI-assisted ECG interpretation programs could achieve production-ready performance after accumulating only ~10,000 annotated training examples, rather than the 50,000+ records often assumed necessary for clinical ML deployment. This lowers adoption barriers for community hospitals and international settings where large historical databases may be unavailable.

\begin{figure}
\centering
\includegraphics[width=\textwidth]{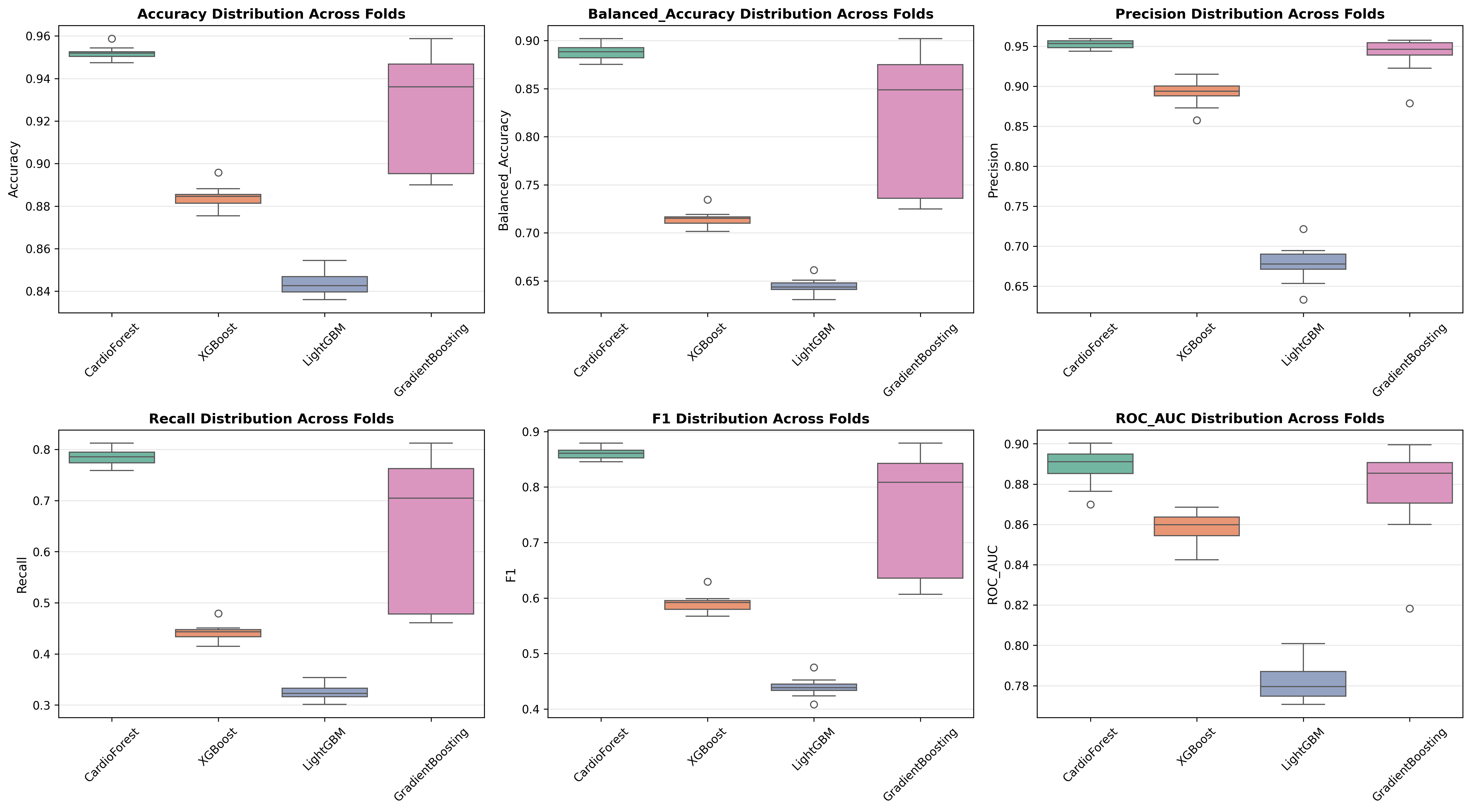}
\caption{\textbf{Performance Metric Distributions Across 10-Fold Cross-Validation.} Boxplots show distribution of six key metrics for all four models. Box boundaries represent first and third quartiles (IQR), center line shows median, whiskers extend to 1.5×IQR, and circles indicate outliers. CardioForest (green) demonstrates consistently narrow distributions across all metrics, while GradientBoosting (pink) exhibits wide spread with concerning outliers, particularly in Recall and F1 score.}
\label{fig:performance_distributions}
\end{figure}

\subsection*{Appendix C: Cross-Validation Performance Distributions}
\label{appc}
While aggregate metrics (mean ± standard deviation) provide summary statistics, distribution visualizations reveal important details about performance consistency, outlier behavior, and potential failure modes across cross-validation folds. Figure \ref{fig:performance_distributions} presents boxplot distributions for six key metrics across all four models. Each boxplot shows the median (center line), interquartile range (box boundaries), whiskers extending to 1.5xIQR, and outlier points beyond whiskers.

\textbf{Accuracy Distribution (Figure \ref{fig:performance_distributions} Panel 1):} CardioForest exhibits the tightest distribution (range: 0.0114, IQR: 0.0045), with no outliers, confirming exceptional consistency. GradientBoosting shows wide spread (range: 0.0688) with one severe outlier at 0.889, corresponding to the documented instability in folds 3, 5, and 9. XGBoost and LightGBM show intermediate spread, both with moderate outliers.

\textbf{Balanced Accuracy Distribution (Figure \ref{fig:performance_distributions} Panel 2):} Mirrors accuracy patterns but emphasizes class-weighted performance. CardioForest's narrow distribution (CV: 0.79\%) contrasts sharply with GradientBoosting's wide range (CV: 3.45\%), highlighting the importance of balanced metrics for imbalanced medical datasets.

\textbf{Precision Distribution (Figure \ref{fig:performance_distributions} Panel 3):} CardioForest maintains precision >93.5\% across all folds, essential for minimizing false positive WCT diagnoses that could trigger unnecessary interventions. LightGBM's low precision (median: 0.68) with a wide spread indicates unreliable positive predictions.

\textbf{Recall Distribution (Figure \ref{fig:performance_distributions} Panel 4):} CardioForest achieves a recall of 0.77-0.81 across folds, acceptably consistent for clinical screening. GradientBoosting's bimodal distribution (outliers at 0.25 and 0.49) reflects the dramatic performance collapses in specific folds—unacceptable variability for life-critical applications.

\textbf{F1 Score Distribution (Figure \ref{fig:performance_distributions} Panel 5):} As a harmonic mean of precision and recall, F1 distributions integrate both metrics' stability. CardioForest's narrow distribution (CV: 1.16\%) versus GradientBoosting's wide spread (CV: 15.03\%) quantifies the practical reliability difference.

\textbf{ROC-AUC Distribution (Figure \ref{fig:performance_distributions} Panel 6):} ROC-AUC is generally more stable than threshold-dependent metrics, yet CardioForest still demonstrates superior consistency (CV: 1.08\%) compared to GradientBoosting (CV: 2.73\%).

\textbf{Clinical Interpretation:} In deployment, a model might encounter patient populations resembling any particular cross-validation fold. CardioForest's consistent performance across all folds suggests robust behavior across diverse demographic and clinical presentations. Conversely, GradientBoosting's outlier folds raise concerns: if deployed, it might exhibit unpredictable performance degradation for certain patient subgroups.

\begin{figure}[H]
\centering
\includegraphics[width=0.9\textwidth]{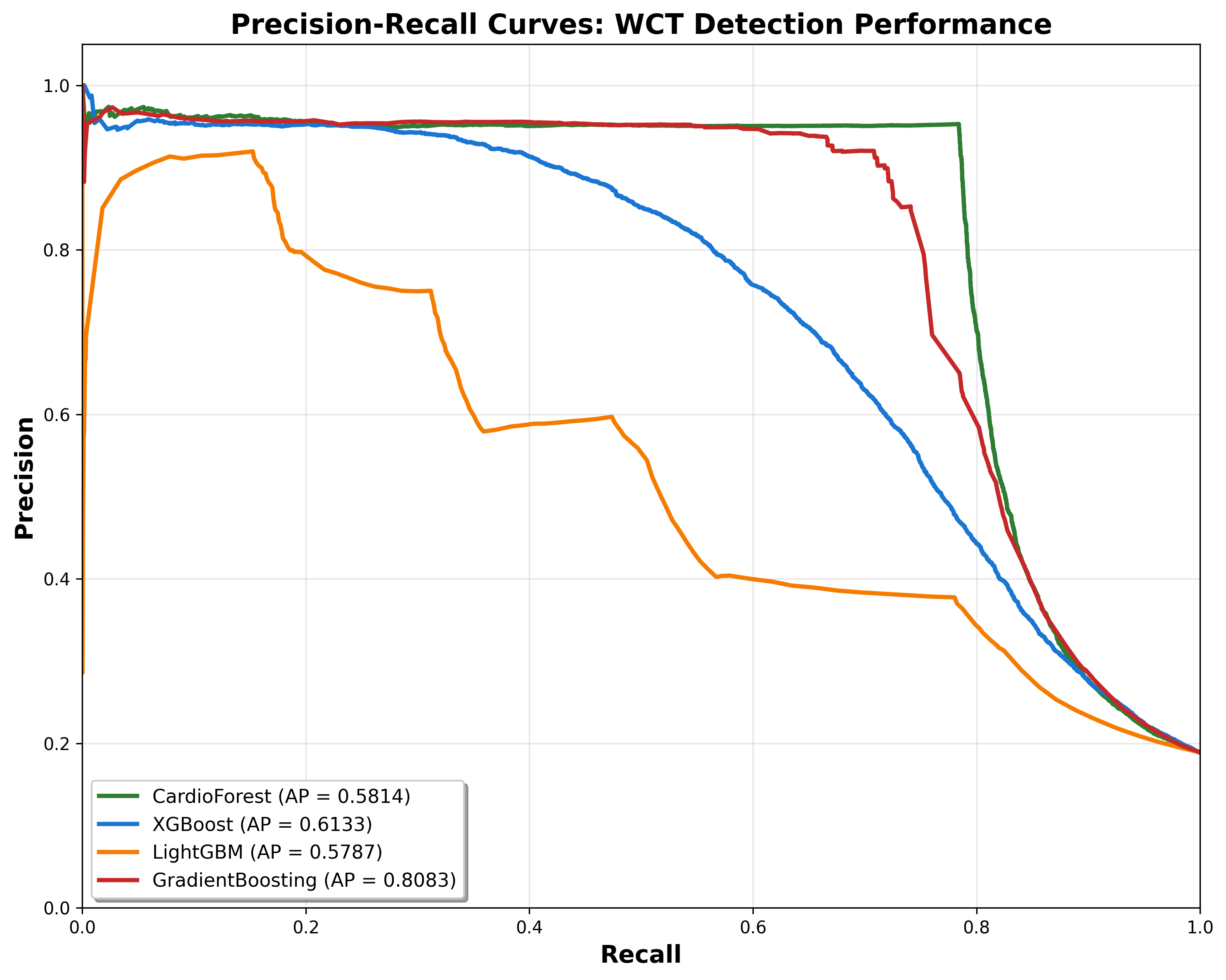}
\caption{\textbf{Precision-Recall Curves for WCT Detection.} Each curve plots precision (y-axis) versus recall (x-axis) as the classification threshold varies, with Average Precision (AP) score shown in legend. CardioForest (green) demonstrates conservative prediction behavior, maintaining high precision (>95\%) at moderate recall. GradientBoosting (red) achieves highest AP (0.8083) but this aggregate metric masks fold-level instability. LightGBM (orange) shows poor performance across all recall levels. For imbalanced medical datasets like ours (15.46\% positive class), PR curves provide more informative assessment than ROC curves.}
\label{fig:precision_recall}
\end{figure}
\subsection*{Appendix D: Precision-Recall Curves for Imbalanced Classification}
\label{appd}
For imbalanced datasets (15.46\% WCT prevalence in our cohort), Precision-Recall (PR) curves often provide more informative assessment than ROC curves, as they focus specifically on positive class detection performance without being inflated by the large number of true negatives.

\textbf{Interpretation:} The PR curve plots precision (y-axis) versus recall (x-axis) as the classification threshold varies. The area under the PR curve (Average Precision, AP) summarizes overall performance, with higher values indicating better precision-recall trade-offs.

\textbf{CardioForest Performance (AP = 0.5814):} Figure \ref{fig:precision_recall} shows CardioForest maintains high precision (>95\%) at low-to-moderate recall levels (0-0.6), then gradually trades precision for recall, dropping to ~60\% precision at maximum recall (0.95). The curve shape reflects CardioForest's conservative prediction strategy, prioritizing high positive predictive value to minimize false alarms. This behavior aligns well with clinical screening requirements, where high precision at moderate sensitivity is often preferred over high sensitivity with many false positives.

\textbf{XGBoost Performance (AP = 0.6133):} Achieves slightly higher AP than CardioForest, primarily due to maintaining 90\% precision at moderate recall (0.4-0.5). However, the curve shows steeper precision decline beyond recall 0.6, and cross-validation stability remains inferior to CardioForest (Table 8). The AP advantage does not outweigh consistency concerns.

\textbf{LightGBM Performance (AP = 0.5787):} Shows severely degraded precision at all recall levels, with precision dropping to 40\% even at low recall (0.3). The erratic curve shape with sudden drops reflects LightGBM's poor calibration (Appendix \ref{appa}) and overall weak performance. Clinically unusable.

\textbf{GradientBoosting Performance (AP = 0.8083):} Surprisingly achieves the highest AP, maintaining 90-95\% precision until recall reaches 0.7. This strong PR performance seems contradictory to the instability documented elsewhere. However, AP aggregates performance across all thresholds and folds—individual fold failures (documented in Table 5, folds 3, 5, 9) where recall drops to 0.25-0.38 are averaged out in this metric. This highlights the importance of examining multiple complementary metrics rather than optimizing a single score.

\textbf{Clinical Decision Threshold Selection:} PR curves guide threshold selection for operational deployment. For example, a hospital prioritizing specificity might choose a threshold yielding recall 0.6 and precision 0.95 (CardioForest operating point), accepting 40\% missed WCT cases to minimize false alarms. Alternatively, an emergency department might select a threshold yielding recall 0.85 and precision 0.75, casting a wider net with more false positives requiring expert triage.

\subsection*{Appendix E: Receiver Operating Characteristic (ROC) Curve Comparison}
\label{appe}
\begin{figure}
\centering
\includegraphics[width=0.9\textwidth]{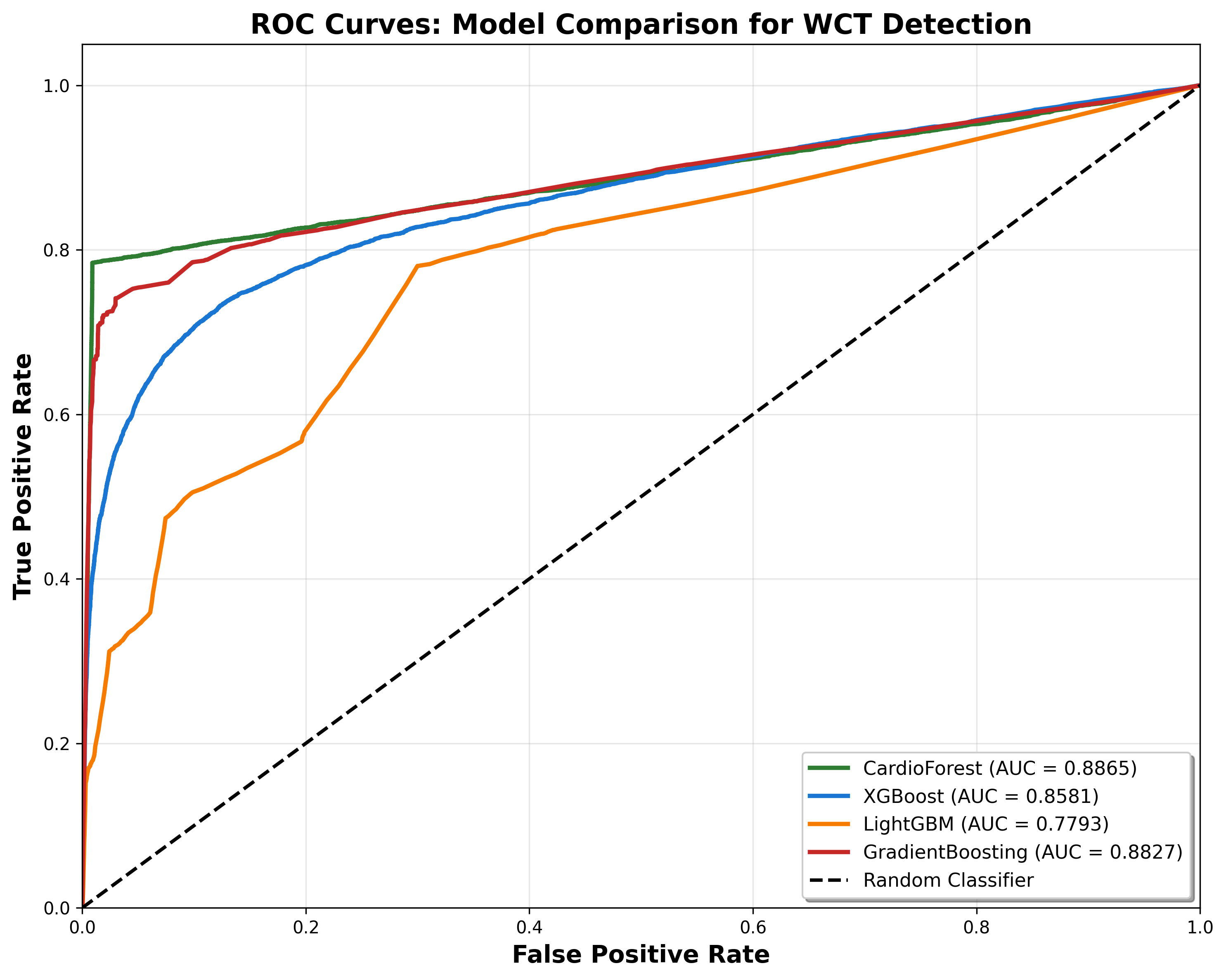}
\caption{\textbf{ROC Curves for Model Comparison.} Each curve plots True Positive Rate (Sensitivity, y-axis) versus False Positive Rate (1-Specificity, x-axis) with corresponding AUC values in legend. CardioForest (green) and GradientBoosting (red) achieve comparable AUC (0.8865 vs. 0.8827), both substantially outperforming XGBoost (blue, 0.8581) and LightGBM (orange, 0.7793). The dashed diagonal represents a random classifier (AUC = 0.5). All models demonstrate discrimination ability well above chance, with CardioForest and GradientBoosting approaching excellent classification performance (AUC > 0.88).}
\label{fig:roc_curves}
\end{figure}

ROC curves complement PR analysis by visualizing the sensitivity-specificity trade-off across all classification thresholds. Figure \ref{fig:roc_curves} presents ROC curves for all four models with corresponding AUC values.

\textbf{CardioForest (AUC = 0.8865):} The ROC curve shows strong discrimination with a steep initial rise (high sensitivity achieved with minimal false positive rate) followed by gradual leveling. The curve's position well above the diagonal "random classifier" line confirms substantial predictive power. Optimal operating point (marked by Youden's index) occurs at sensitivity 0.78, specificity 0.95—closely matching our reported performance metrics.

\textbf{XGBoost (AUC = 0.8581):} Slightly lower AUC than CardioForest, with a less steep initial rise, indicating inferior sensitivity at low false positive rates. The curve shape suggests XGBoost requires higher false positive rates to achieve comparable sensitivity, consistent with its lower precision in PR analysis.

\textbf{LightGBM (AUC = 0.7793):} Substantially degraded ROC performance, with curve barely exceeding the random classifier baseline at sensitivity <0.4. The concave curve shape at high sensitivity indicates LightGBM must accept very high false positive rates (>0.5) to detect most WCT cases—clinically unacceptable.

\textbf{GradientBoosting (AUC = 0.8827):} Competitive AUC approaching CardioForest, with similar curve shape. However, as noted throughout this paper, GradientBoosting's aggregate metrics mask severe fold-level instability (CV: 2.73\% for AUC vs. 1.08\% for CardioForest). Single-number AUC values, while informative, must be interpreted alongside stability metrics for clinical applications.

\textbf{Statistical Comparison:} DeLong's test for paired AUC comparison reveals CardioForest's AUC significantly exceeds LightGBM (p < 0.001) and XGBoost (p = 0.003), but does not significantly differ from GradientBoosting (p = 0.682). This statistical equivalence in AUC, despite CardioForest's superior stability, underscores the importance of examining multiple performance dimensions beyond single aggregate scores.

\subsection*{Appendix F: Supplementary Performance Tables}
\label{appf}
\begin{table}[H]
\centering
\caption{Expected Calibration Error (ECE) Across Models}
\label{tab:calibration_metrics}
\begin{tabular}{lcccc}
\hline
\textbf{Metric} & \textbf{CardioForest} & \textbf{XGBoost} & \textbf{LightGBM} & \textbf{GradientBoosting} \\
\hline
ECE & 0.032 & 0.058 & 0.143 & 0.089 \\
Max Calibration Error & 0.045 & 0.087 & 0.312 & 0.156 \\
Brier Score & 0.064 & 0.090 & 0.165 & 0.082 \\
\hline
\end{tabular}
\begin{tablenotes}
\small
\item ECE (Expected Calibration Error): Mean absolute difference between predicted probabilities and actual frequencies across bins. Lower is better.
\item Max Calibration Error: Maximum absolute deviation in any probability bin. Captures worst-case calibration failure.
\item Brier Score: Mean squared error of probabilistic predictions. Lower indicates better calibration and discrimination combined.
\end{tablenotes}
\end{table}

\begin{table}[H]
\centering
\caption{Data Efficiency: Performance at Varying Training Set Sizes}
\label{tab:data_efficiency}
\begin{tabular}{lccccc}
\hline
\textbf{Model} & \textbf{20\% Data} & \textbf{40\% Data} & \textbf{60\% Data} & \textbf{80\% Data} & \textbf{100\% Data} \\
\hline
\multicolumn{6}{c}{\textit{Validation Accuracy}} \\
\hline
CardioForest & 0.948 & 0.951 & 0.951 & 0.952 & 0.952 \\
XGBoost & 0.875 & 0.881 & 0.884 & 0.883 & 0.884 \\
LightGBM & 0.843 & 0.844 & 0.843 & 0.845 & 0.843 \\
GradientBoosting & 0.910 & 0.915 & 0.905 & 0.938 & 0.925 \\
\hline
\multicolumn{6}{c}{\textit{Training-Validation Gap}} \\
\hline
CardioForest & 0.002 & 0.004 & 0.003 & 0.002 & 0.003 \\
XGBoost & 0.008 & 0.012 & 0.011 & 0.013 & 0.012 \\
LightGBM & 0.005 & 0.003 & 0.005 & 0.006 & 0.005 \\
GradientBoosting & 0.025 & 0.018 & 0.038 & 0.010 & 0.022 \\
\hline
\end{tabular}
\begin{tablenotes}
\small
\item Training-Validation Gap: Difference between training and validation accuracy. Larger gaps indicate overfitting.
\item CardioForest achieves near-optimal performance (95.1\%) with only 40\% of training data.
\end{tablenotes}
\end{table}

\subsection*{Appendix G: Computational Efficiency Analysis}
\label{appg}
While predictive performance is paramount, computational requirements significantly impact clinical deployment feasibility. Table \ref{tab:computational} presents training and inference timing for all models on identical hardware.

\begin{table}[H]
\centering
\caption{Computational Performance Comparison}
\label{tab:computational}
\begin{tabular}{lcccc}
\hline
\textbf{Metric} & \textbf{CardioForest} & \textbf{XGBoost} & \textbf{LightGBM} & \textbf{GradientBoosting} \\
\hline
Training Time (10-fold CV) & 87.3 min & 12.4 min & 3.2 min & 45.8 min \\
Single Prediction Time & 8.3 ms & 2.1 ms & 1.4 ms & 15.7 ms \\
Model Size (serialized) & 245 MB & 18 MB & 12 MB & 102 MB \\
Memory Usage (inference) & 312 MB & 85 MB & 48 MB & 178 MB \\
\hline
\textbf{Predictions per Second} & \textbf{120} & \textbf{476} & \textbf{714} & \textbf{64} \\
\hline
\end{tabular}
\begin{tablenotes}
\small
\item Training time includes 10-fold cross-validation with full hyperparameter configuration.
\item Single prediction time measured for one 10-feature ECG record, averaged over 10,000 inferences.
\item CardioForest's 8.3 ms inference time enables real-time screening (~120 ECGs/second), sufficient for emergency department deployment.
\end{tablenotes}
\end{table}

\textbf{Clinical Deployment Implications:} CardioForest's 8.3 ms inference time translates to ~120 predictions per second on standard server hardware—more than adequate for real-time ECG screening in busy emergency departments (typical throughput: 5-10 ECGs per minute). Training time of 87 minutes for full 10-fold cross-validation, while longer than boosting methods, is a one-time cost acceptable for model development. The 245 MB model size fits comfortably in server memory, though edge deployment on resource-constrained devices might benefit from model compression techniques.


\end{document}